# An Integrated Framework Integrating Monte Carlo Tree Search and Supervised Learning for Train Timetabling Problem

Feiyu Yang [a]

[a] School of Information Science and Technology, Southwest Jiaotong University, Chengdu 611756, China

**Abstract** The single-track railway train timetabling problem (TTP) is an important and complex problem. This article proposes an integrated Monte Carlo Tree Search (MCTS) computing framework that combines heuristic methods, unsupervised learning methods, and supervised learning methods for solving TTP in discrete action spaces. This article first describes the mathematical model and simulation system dynamics of TTP, analyzes the characteristics of the solution from the perspective of MCTS, and proposes some heuristic methods to improve MCTS. This article considers these methods as planners in the proposed framework. Secondly, this article utilizes deep convolutional neural networks to approximate the value of nodes and further applies them to the MCTS search process, referred to as learners. The experiment shows that the proposed heuristic MCTS method is beneficial for solving TTP; The algorithm framework that integrates planners and learners can improve the data efficiency of solving TTP; The proposed method provides a new paradigm for solving TTP.

## I. Introduction

The train timetabling problem has been extensively studied in the fields of transportation, operations research, and computing science due to the criticality of its functions, the extensiveness of its applications, and the complexity of its computation. The research on the TTP, at the current stage, is mainly focusing on the double-track railways with stronger infrastructure capacity, such as the TTP optimization of high-speed railways or rail transit, to meet the needs of passenger transportation [1][2][3][4]. Many countries in the world (such as Australia[5], USA[6] and China[7]) still use a large number of single-track railways in the aspects of serving the transportation demand for energy commodities and other bulk commodities, which is helpful for maintaining energy security[6][8]. However, the infrastructure capacity of single-track railways is more limited than that of double-track railways, making it more difficult to generate their optimal timetables. Theoretically speaking, the TTP of single-track railways is a typical strong coupling and large-scale NP-hard problem[9][10], and how to solve this problem effectively has always been a challenge.

At present, the popular methods for settling the TTP can be roughly divided into exact planning, heuristics, and reinforcement learning (RL). For the exact method, it is hard to use it to solve the TTP when there is a large-scale problem. Fast heuristic methods are difficult to balance between the quality of solutions and its required computing cost, and developing efficient heuristic rules is not easy, either. The RL method have recently become a new hotspot in the research on TTP, and its core is to iteratively increase the probability of being selected for actions that can achieve good performance. However, it is difficult to obtain feasible solutions of the TTP with positive rewards when it comes to complex problems, which will lead to no effective rewards for the RL method based on random exploration. More than that, building a Markov environment in solving the TTP of single-track railways is not easy thing, especially when the additional time of the acceleration and deceleration of trains is taken into account, because the state of the environment depends on the state of the next action (the decision for a train to halt at or depart from the next station), which makes a Markov environment hard to achieve. Another difficulty encountered by the RL method is that the applicability of different TTP instances is not extensive enough. RL algorithms highly rest on the sampling information provided by problem instances, so, generally speaking, they need to be retrained to adapt to the new instances even if there is a very slight change in problem instances, such as a change in the train arrival time. In addition to the above methods, some artificial intelligence methods in other fields tend to pursue generalization to some extent, meaning that the algorithms would also give reasonable output inferences when confronted with untrained data.

Since the vanilla MCTS algorithm[11] was proposed, its variants have rapidly emerged. In this paper, the MCTS framework refers to the algorithms based on following four phases: *Selection*, *Expansion*, *Simulation*, and *Back Propagation*. At first, the goal of the vanilla MCTS framework is using the tree search to improve the accuracy of Monte Carlo method[11], this method has outperformed in 9x9 Go programs. Then, the introduction of the idea Bandit makes the vanilla MCTS evolve into the Upper Confidence Bound Apply to Tree (UCT), an improved MCTS algorithm gaining widespread attention [12]. The UCT has been proved to be effective and has extraordinary potential in the AlphaGo, a Go decision-making program [13], and it has been used and extended into operations research such as Mixed-integer Programming and scheduling issues as well [14][15]. Therefore, we propose a method that using UCT to solve TTP for single-track railways and two improved algorithms based on the UCT, UCT_MAX1 and UCT_MAX2, as shown on the left side of Fig. 1. The three algorithms, UCT, UCT_MAX1 and UCT_MAX2, are collectively called the general UCT algorithms, because they have similar functions and algorithmic structures in the proposed method. Like the general RL method, the solutions of the general UCT algorithms are asymptotically optimal when the computing budget is optimal. However, in the light of the limited computing budget in the real world, we put forward a simple heuristic strategy Conflict-free Steps As a Value (CSAV) for the general UCT algorithms to enhance their searching efficiencies without affecting their own asymptotic optimality. From the perspective of the RL, the MCTS algorithms (the general UCT algorithms with or without the CSAV) for solving TTP proposed in this paper are regarded as a model-based planner [16], and their problem-solving abilities will be well demonstrated. Moreover, we also propose a supervised learning system as a learner based on deep convolutional neural networks (CNN), as shown on the right side of Fig. 1. The learner adopts the labeled state data generated by the planner for training, infers the new input states by taking the trained deep neural network as a value approximate function (VAF), and outputs the corresponding estimated values. The method presented in this paper has a combination of several optional algorithms, so it is regarded as an algorithm framework.

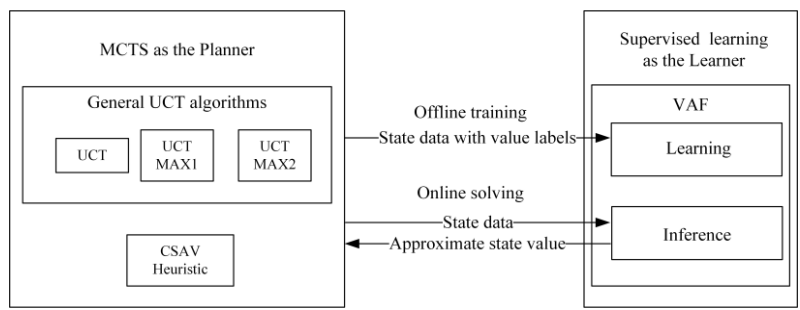

Fig. 1 A diagram of the proposed method that integrates MCTS planning and supervised learning for solving TTP

Different from conventional planning or supervised learning methods, the planner can not only settle the TTP of single-track railways, but also provide a large amount of labeled training data for supervised learning, but its disadvantage is that it requires a long searching time and does not have generalization capability. On the contrary, the learner, with the generalization ability, is capable of extracting the features of massive training data and learning its commonness, and the inference time is so short as to be not worth considering. By embedding the inference process of the offline training learner into the online searching process of the planner, this integrated method is more accurate. From this point of view, the planner and learner in the proposed framework are collaborative and complementary.

## A. Literature review of the TTP

The TTP has been intensively studied since it was modeled as a Mixed Integer Programming (MIP) by Szpigel in 1973[17]. In terms of its infrastructure, the TTP mainly comes from single-track railways [18][19][20][21][22][23][24] and double-track railways[1-2]. Each open track of a double-track railway is unidirectional, and trains traveling in each direction can quickly pass each other, such as high-speed railways and rail transit. While for a single-track railway, trains in both directions share the same track, which limits its capacity. When a train is running at a section, the rest of the trains trying to pass the section can only wait on the station tracks, that is to say, the trains that need to cross and overtake can only wait at the station. In recent studies, the track allocation in railway stations and TTP are jointly optimized to improve the efficiency of railway infrastructure [25][26][27][28][29]. The TTP can be divided into two parts, train planning and train operations, from the management point of view [1][10][30]. Train planning involves the determination of train timetables weeks or months in advance; and train operations generally focus on the timetable recovery when there is interference, which depends on the predetermined timetables. As for fixed train timetables, periodic and non-periodic timetables fall into the category of TTP [31], and periodic timetables will inevitably lose part of the line capacity [30]. For more details, the outstanding surveys of TTP are what we recommend to the interested readers [1][3][10][2][32][33][34][35].

In this paper, the non-periodic TTP of single-track railways in train planning stage is studied, and the number of valid tracks in stations and the additional time of acceleration and deceleration during train dwelling at stations is also considered. Our goal is to find a conflict-free single-line train timetable with the minimal total running time, which means that we need to ensure that constraints are met while minimizing the train running time. Its core is that when there are many trains running in a certain time range, there are a large number of space-time constraints associated with each train trajectory in the large solution space. Moreover, the distribution of various constraints changes dynamically with the decisions on train timetabling. In this regard, considering the large scale and high coupling of such problems, it is difficult to obtain enough feasible solutions. Thus, an efficient algorithm is needed to achieve this goal while searching for feasible solutions.

TTP algorithms are mainly divided into exact, heuristic and simulation-based algorithms. For exact algorithms, branch and bound and commercial solvers are often used for solving TTP, and the advantage of the algorithms is that the optimal solutions can be found. Petering, Matthew E. H et al. [36] propose a mixed-integer linear programming (MILP) model with the combination of the TTP and platforming problem, which is also helpful for estimating the capacity of the lines, and they use CPLEX to simultaneously minimize the scheduling cycle length and the total travel time of all trains. Yu-Jun Zheng defines a "headroom" function for computing the maximum number of emergency trains that can be inserted between two adjacent regular trains without any disturbance, from which a linear algorithm for finding ideal solutions is derived based on a non-linear programming model; and the quasi-ideal and feasible solutions can be found based on the ideal solutions on a set of alternative timetables [37]. Chuntian Zhang et al. point out a MILP model considering the passengers' expected departure interval to minimize the operation cost and total travel time of trains simultaneously. In their study, the bi-objective problem is transformed into a single-objective problem by linear weighted methods, and the problem is solved with GUROBI[38]. However, due to the complexity of TTP, it is hard for exact algorithms to find the optimal solution as the size of TTP instances increases, so many researchers choose heuristic algorithms to solve complex TTP. Valentina et al. conduct a study on the TTP for a railway node in order to obtain conflict-free and adjusted timetables that deviate from the original timetables, considering that multiple railway operators in Italy may propose their own train timetables and cause conflicts. For this purpose, an integer linear programming model is established and an iterative heuristic algorithm is adopted [30]. Yin YH et al. combine the features of periodic timetabling with those of demand-responsive timetabling design for a bidirectional rail transit line considering rolling stock circulation. Also, an MIP model with the objective of minimizing total passenger waiting time and weighted total train travel time is formulated and a three-phase heuristic algorithm is developed [39]. Jungang Shi et al. present an integer linear programming model for collaboratively optimizing the train timetables and accurate passenger flow control strategies on an oversaturated metro line, and a hybrid algorithm, which combines an improved local search and CPLEX solver, is designed to search for high-quality solutions[40]. Qing Zhang et al. provide us a method involving a commercial solver and heuristic algorithm, and formulate a binary integer programming model for dealing with the TTP, platforming, and infrastructure maintenance scheduling[29]. For heuristic algorithms, optimization models are used to describe the objectives and constraints of problems, and the variables of the models are solved by heuristic rules. Besides, the decision-making process is generally independent of the dynamic train operation mechanisms within the system. However, on one hand, the heuristic algorithms are highly relying on the features of problems, so their generalization is limited, and it is difficult to design hand-crafted heuristic rules with excellent performance. On the other hand, the feature information that can support more relevant heuristic rules could be lost because of the lack of descriptions on dynamic train operation mechanisms.

Unlike some heuristic methods relying on optimization models, simulation-based heuristic or machine learning methods, especially RL methods, have triggered heated discussions in the TTP research and other related fields[24][32][41][42][43][44][45][46]. Simulation-based methods use optimization models for quality evaluation instead of rule formulation, or adopt dynamic train operation mechanisms to design the process inside the simulation system. For example, Xu, Xiaoming et al. propose a simulation-based method to assign locomotives and train timetabling simultaneously, and evaluate the quality of the generated solutions by taking the objective value as a valid lower bound [47]. Furthermore, since the simulation-based methods usually make sequential decisions on trains, many decision trajectories including system states can be generated during the decision-making process, thereby being used to support supervised or unsupervised learning in learning algorithms. Khadilkar Harshad offers a scalable tabular RL method to obtain train timetables by recording train trajectories generated with a discrete event simulation system[20]. The RL algorithm itself is not interested in various types of train operational constraints, but the update rules of the simulation system characterize them, reflecting a benefit of simulation-based methods. Although the tabular RL method is simple and intuitive, it is unlikely to store such a large-scale state space when the state space of problems increases, so the non-linear deep neural network is widely used to approximate the value of the state, which is also known as deep reinforcement learning[16]. For example, Li and Ni come up with a multi-agent deep reinforcement learning (MADRL) algorithm based on the Actor-critic architecture to solve the action strategies of trains. In their work, they treat each train as an agent, and every agent has its own strategy for interacting with the environment and is subject to the constraints of a centralized strategy at the time of training. Since the MADRL algorithm does not take the capacity of a station into account, it is not applicable to the scenario considered in this paper [7].

The prevailing RL algorithms, which are model-free, have excellent performance in solving the TTP, but they have limitations in solving the large-scale one. The main reason for that phenomenon is that there are two drawbacks in the bootstrapping data acquisition method when tackling with the large-scale TTP. Firstly, it is tough for algorithms to search for the solutions that are feasible enough by random exploration. In other words, when the algorithms are unable to search for positive experiences (how to obtain feasible solutions) to learn, it would degenerate into the breadth-first traversal algorithm, thereby failing to search for positive experiences in the large solution space again. The second reason is that once a feasible solution is obtained by the RL algorithm, on the account of the complex coupling of the TTP, the algorithm would continuously reinforce its elements, rather than easily jumping out of the local feasible solutions through random exploration. A slight change in the elements of the solution may result in an infeasible solution. Theoretically, it is always assumed that local solutions can be avoided through nearly infinite exploration, but, practically, it is clearly inefficient. Here, the model-based RL algorithm can help to resolve the dilemma, improve the efficiency of data use, and maintain a balance of both the Exploration and Exploitation [16]. The MCTS framework is the most popular planning component of model-based RL algorithms. It can be highly extended and integrated with learning and heuristic algorithms in various ways, with the success of AlphaGo as the best example for demonstrating the potential of the combination [13]. In summary, the MCTS framework is a promising way to study the TTP, especially when it comes to combining with other algorithms.

### B. Literature review of MCTS Framework

There are a few reports on solving the TTP with MCTS, but the deformation, expansion, and application of this framework in other fields are increasing rapidly. For more details, we recommend readers reading excellent review studies[14][15]. In the field related to the TTP, Rongsheng Wang et al. propose a train rescheduling method based on MCTS, which selects an optimal train departure sequence through the data structure node of the tree to reduce the average train delay on a railway line[44]. From its simulation results, it can be demonstrated that the proposed method can bring about a rescheduling strategy of a less average total delay within a faster computing speed, which is superior to the conventional methods. However, it is not applicable to generate an initial train timetable, because this method is developed on the basis of a pre-specified train timetable.

Recently, many researches have indicated that the MCTS is highly extensible and easy to integrate. Sylvain Gelly and David Silver show us a fast action-value estimation method with MCTS in their work[48]. The total success probability of each action in the *Simulation* process is regarded as a factor to measure the quality of the action. Although this estimation method does not accurately reflect the real value expectation of the action, it still improves the performance of the algorithm in Go. In addition, the study points out that the best performance of an action that can be achieved (the max operator) is used to measure the value when evaluating an action, but the average operator (or the mean operator) is more robust in Go, because Go does not have an "ideal action distribution". Things are completely different for the TTP. If there is no conflict, each train would run separately with the minimum running time, and the total train travel time is minimal as well. Thus, we adopt the max operator in a reasonable manner to measure the values of actions in TTP. Nasser R. et al. and Shahriar A. et al. put forward a hyper heuristic method based on the MCTS framework, respectively, illustrating the fine integration of the MCTS and heuristic algorithms, and the favorable performance of the integrated methods for dealing with the problems in combinatorial optimization and resource-constrained timetabling [49][50]. Given that the vanilla MCTS algorithm faces the same difficulty as the RL algorithms, which is that it is hard to search for feasible solutions with the growing scale of TTP instances, we go about strengthening the solving ability by embedding heuristic policies into the MCTS framework.

The combination of the MCTS framework with learning algorithms, nowadays, seems to be more favored by researchers than with heuristic algorithms, which can be well demonstrated by a series of successful applications in the field of Go. Although there is no literature report on a general integration paradigm of the MCTS framework and learning algorithms, the popular integration can be roughly divided into two types. One is using the MCTS framework to improve the efficiency of data use and the diversity of data for other learning systems, such as RL, and the other is to construct a learning system for the MCTS framework, such as using deep neural networks to enhance the generalization and accuracy of the MCTS framework itself. Here are some specific examples. Gelly and Silver consider a combination of three methods to learn online and offline knowledge, and then the learned knowledge is used for the *Simulation* policy, value evaluation, and prior knowledge of the MCTS framework, respectively, by taking the TD ($\lambda$) algorithm as the learner [51]. Their attempt lays a foundation for the success of AlphaGo. In power grid energy dispatching, Yuwei Shang et al. regard the Q-learning algorithm as a learner and the MCTS framework as a data enhancement method to facilitate the efficiency of data use in the learning process[52]. Chenjun Xiao et al. store states into memory in the MCTS planning process, and update the values corresponding to the states and surrounding values, according to the generalized distances between states[53]. Their purpose is to generalize state values based on the assumption that the state values close to each other are similar. As a matter of fact, it is difficult to define the distances between states for values, especially for issues such as the TTP, because even a subtle difference between states could bring about a huge difference in the state values. DouZero [54] is a very popular card game application in China, and it involves a deep MCTS algorithm. The idea of the algorithm is to directly approximate the states and values obtained by the MCTS with deep neural networks. Thanks to the high nonlinearity of the deep neural networks, the algorithm is capable of achieving the state-of-the-art performance. Directly learning the data in the MCTS process in comparison with learning with the RL can avoid the repeated iterative updates and overestimation of state values [16], such as the Q-learning algorithm. In addition, if the MCTS algorithm is counted as a data generator (a dataset) and a state evaluator (a label), this learning mode can be generalized to a paradigm similar to the supervised learning, as it is in computer vision. Based on this mode, state values are able to be generalized for a class of problem instances instead of estimating each state value in a specific problem instance, which may overcome the limitations of the current RL algorithm up to a point. Therefore, we manage to directly generalize states and their estimated values generated by the MCTS framework with a Convolutional Neural Network (CNN), and integrate the network into the MCTS framework as an independent component.

## C. Challenges and Contributions

Considering the pros and cons of the various methods summarized in the literature review above, the goal of this paper is to adopt the MCTS and supervised learning technique to present a TTP solving method integrating with a planner and a learner for single-track railways. Therefore, the challenges that we have faced are reflected in the following questions, and our contributions are described in the answers to each question:

Q1: How do MCTS algorithms solve TTP of single-track railways and serve as a planner in the proposed method?

- Based on the MIP model of TTP for single-track railways, a state matrix that can describe train running trajectories and their corresponding constraint distribution in time and space dimensions is shown. With this state matrix, a dynamics model of train operation simulation is set up, and then a train operation simulator on the strength of the dynamics model is developed.
- The state matrix in the simulator is taken as the state of a MCTS tree node, and the train dwell time, the decision variable, is treated as the path of the tree. The MCTS tree is constructed and solved using UCT algorithm. To make the UCT algorithm more applicable for solving TTP, the general improved algorithms UCT_MAX1 and UCT_MAX2 are presented. More than that, a heuristic strategy CSAV is also demonstrated to enhance the searching ability when computing resources are insufficient.
- Based on the MIP model and dynamics model of the simulator, the algorithms, combined a UCT, UCT_MAX1, or UXT_MAX2 which is used for solving TTP with the heuristic strategy CSAV, are considered as a planner. The planner solves TTP instances for single-track railways, and takes the state data collected during the computing process and the corresponding state values (state-value pairs) as the training datasets for the supervised learning.

Q2: How to design a supervised learning system and how does it function as a learner to learn data?

- Taking the data preprocessing method from AlphaGo for reference, the dynamic history of train trajectories and the time-space distribution of constraints are obtained by stacking the state data matrix in datasets provided by a planner. Inspired by attention mechanisms, the region representing the nearest train movement in the training data is clipped in the data preprocessing, and the clipped data together with the full state data constitutes two inputs of the deep CNN.
- In this paper, a deep convolutional network with multiple input processing paths is put forward to extract the features of state data and learn the commonalities of these data. Similar to the supervised learning in other fields, the network has a certain degree of data generalization ability after training. That is to say, when the untrained state data are input, the network can output the approximate values of the corresponding state values. The network is regarded as a non-linear VAF. The state data learned by VAF are the matrix representation of train timetables, so the train timetables are generalized, to some extent.

Q3: How is the proposed method formed with the combination of a planner and a learner?

The train timetables mentioned in this paper are at the planning level. To distinguish the training and solving process of the proposed method, here are some explanations. The process of actually computing and solving the TTP instances is called online solving; and the rest of the process, the process of solving the manually selected or randomly generated TTP instances by a planner to obtain a training dataset and subsequently learning a large amount of data with a learner is called offline training.

- Offline training: For some specific timetabling scenarios, the planner can adopt one or more of the proposed MCTS algorithms to solve manually set or randomly generated TTP instances. That is, according to the given train running scheme and route infrastructure information, the planner can get the corresponding train timetables and store the state data and its value data generated in the computing process. On this basis, the VAF of the learner uses the supervised learning to update the weights in the neural networks, so as to minimize the deviation between the approximate values output from the networks and the values estimated by the planner. The experimental results suggest that the proposed training method of combining a planner and a learner is effective and saves the cost of data labeling in general supervised learning methods.
- Online solving: We offer a method which uses VAF to estimate the values of states encountered by an algorithm in a planner, denoted as UCT_VAF. The method utilizes the Warm up technique to provide approximate values of states while the MCTS algorithm is solving, thus guiding the MCTS algorithm to allocate computing resources in a more reasonable manner so that promising tree nodes are evaluated in more simulations. The experimental results show that this method improves the quality of the planner in finding solutions.

In brief, firstly, a novel train operation simulation method is presented in this paper. Secondly, based on this simulation method, a MCTS algorithm, UCT, for solving TTP of single-track railways is proposed and improved, and a deep CNN is designed to learn the knowledge obtained by the MCTS algorithm. Thirdly, the deep CNN is combined with the MCTS algorithm to improve the searching ability. The framework proposed by us can be highly extended, so it provides a new perspective of combining planning and learning methods for the research on the TTP for single-track railways and related fields.

## II. Problem Statement, Models and Background

This section first describes the TTP of single-track railways and gives a MIP optimization model. Next, a simulator based on the matrix-form state is proposed, and the dynamics model of the simulator is given. Then, this section describes the general forms and algorithms of the Vanilla MCTS and UCT as the background of Section III. Table 1 provides a description of the symbols used in this section.

Table 1. Symbols defined for the TTP models, Vanilla MCTS, and UCT algorithm

| Symbol | Description |
|---|---|
| $i$ | $i$ denotes a train. $i \in I$, $I \subset \mathbb{N}^+$ represents the sequence consisting of all trains, and the train sequence is sorted according to the first departure time of each train. |
| $I(i)$ | $I(i) \subset I$, the image of mapping $I(i)$ is a subsequence of $I$, indicates trains in same direction to train $i$. |
| $j$ | $j$ indicates a station, $j \in J$, $J \subset \mathbb{N}^+$ represents the sequence consisting of all stations, and the station sequence is sorted according to the direction from the departure station to the destination station. |
| $J(i)$ | $J(i) \subseteq J$, the image of mapping $J(i)$ represents the sequence of all stations between the departure station and the destination station of train $i$, which is a subsequence of $J$. |
| $J(i)_k$ | The $k$th element of the station sequence $J(i)$, $k \in [1, |J(i)|]$. |
| $j(i)$ | $j(i) \in J(i)$, an image of the mapping $j(i)$ represents the next station of the station where train $i$ is located according to the running direction of $i$. |
| $s$ | $s$ indicates a section, $s \in S$, $S \subset \mathbb{N}^+$ represents the sequence consisting of all |

| | sections, the sequence is sorted according to the direction of $J$, $|S| = |J| - 1$. |
|---|---|
| $s(j,j')$ | $s(j,j') \in S$, an image of mapping $s(j,j')$ indicates the section between the station $j$ and its adjacent station $j'$. |
| $v_{i,j}$ | A nonnegative integer variable that represents the dwell time of train $i$ at the station $j$. |
| $V$ | After the train route planning, the first departure time, departure stations, and destination stations of all trains have been specified, and then we can define a dwell time decision variable sequence $V = [v_{1,J(1)_1}, v_{1,J(1)_2}, \dots, v_{1,J(1)_{|J(1)|}}, v_{2,J(2)_1}, \dots, v_{|I|,J(|I|)_{|J(|I|)|}}]$ for all trains according to the order of the first departure time to determine train timetables. |
| $e_{i,j}$ | Arrival time of train $i$ entering station $j$. |
| $d_{i,j}$ | Departure time of train $i$ from station $j$. |
| $t_{i,j}^{acc}$ | Time of acceleration when train $i$ departs from station $j$. |
| $t_{i,j}^{dec}$ | Time of deceleration when train $i$ arrives at station $j$. |
| $b_{i,j}$ | A binary decision variable. $b_{i,j}$ is 1 when train $i$ needs to dwell at station $j$ and it is 0 when train $i$ passes station $j$ directly. |
| $\tau_{no}$ | Time interval between any two trains arriving at a same station. |
| $\tau_d$ | The interval between the time when a train departs from a section and arrives at a station and the time when another train departs from the station and then running into the same section. |
| $V_{max}$ | Maximum train dwell time at a station. |
| $T_{max}$ | Maximum timetabling time (it starts with 0 in this paper). |
| $t_{i,s}^{nom}$ | The nominal running time of train $i$ through section $s$ is specified in advance by a series of experiments, such as traction power experiments. In this paper, the nominal running time is simplified as the ratio of the section length to the train speed. On this basis, we have considered the impact of the acceleration and deceleration process of the train on the actual train running time as well. |
| $b_{i,j,t}^{station}$ | A binary variable. At time $t$, if train $i$ dwells at station $j$, $b_{i,j,t}^{station}$ is 1, otherwise it is 0. |
| $b_{i,s,t}^{section}$ | A binary variable. At time $t$, if train $i$ dwells at section $s$, $b_{i,s,t}^{section}$ is 1, otherwise it is 0. |
| $C_j$ | A constant that represents the number of parallel tracks of station $j$. |
| $f_{obj}$ | The objective function. |

## A TTP

1) The mesoscopic infrastructure

In this paper, the infrastructure of single-track railways is modeled at the Mesoscopic level [55][56]. As shown in Fig. 2, a railway line consists of several stations and sections, and each station may have multiple parallel tracks, but each section has only one track. Limited by signal facilities on single-track railways, normally, only one train can run in each section at the same time. It is assumed that a track is occupied by at most one train at the same time. If a station track can carry more than one train, the track can be simplified as multiple parallel tracks. Compared with the section length, the length of the station is shorter. The length of both station tracks and trains are not considered. If a train dwells at the station, the additional time specified for acceleration and deceleration will be added to the dwell time.

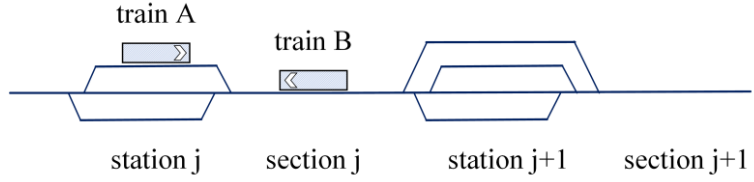

**Fig. 2.** A layout example of single-track railways

2) The MIP model

Our objective is to minimize the total conflict-free travel time for all trains, as shown in Eq. 1. The decision variable $v_{i,j}$ is the dwell time of each train at each station. The binary intermediate variable $b_{i,j}$ is introduced to indicate whether the acceleration and deceleration process has been performed, and the other two binary intermediate variables $b_{i,j,t}^{station}$ and $b_{i,s,t}^{section}$ are presented to express the capacity of the railway line infrastructure. The constraints of the model are divided into headway constraints, running time constraints and constraints of the railway line infrastructure capacity. The MIP model is as follows:

$$f_{obj} = \min \sum_{i \in I}(d_{i,J(i)_{|J(i)|}} - e_{i,J(i)_1}) \quad (1)$$

$$s.t. \begin{cases} e_{i,j} \geq 0 \quad (2) \\ d_{i,j} \leq T_{max} \quad (3) \\ d_{i,j} \equiv v_{i,j} + e_{i,j} \quad (4) \\ v_{i,j} \leq V_{max} \quad (5) \\ |e_{i,j} - e_{i' \in I-i,j}| > \tau_{no} \quad (6) \\ |d_{i,j} - e_{i' \in I-I(i),j}| > \tau_d \quad (7) \\ e_{i,j(i)} \equiv d_{i,j} + t_{i,s(j,j(i))}^{nom} + b_{i,j}t_{i,j}^{acc} + b_{i,j(i)}t_{i,j(i)}^{dec} \quad (8) \\ \sum_{i \in I} b_{i,j,t}^{station} \leq C_j, \forall j \in J \text{ at } \forall t \in [0, T_{max}] \quad (9) \\ \sum_{i \in I} b_{i,s,t}^{section} \leq 1, \forall s \in S \text{ at} \forall t \in [0, T_{max}]. \quad (10) \end{cases}$$

The travel time of each train is calculated according to the departure time and arrival time of the train, including the dwell time and the time for acceleration and deceleration. The running time of a train passing through a section described in Eq.8 depends not only on the time when the train enters the section, but also on whether the train dwells at the current station and next station. Four different situations are expressed in Fig. 3. Eq.2, Eq.3, and Eq.5 represent the first departure time limit of trains, the maximum train running time limit, and the maximum train dwell time limit at a station, respectively. Eq.4 and Eq.8 describe the processes of a train stopping at a station and running in a section, while Eq.9 and Eq.10 limit the number of tracks in stations and sections. In order to ensure the safety of railway operations, it is necessary to reserve sufficient preparation operation time for the railway line infrastructure. Therefore, Eq.6 denotes the interval between any two trains arriving at a station, and Eq. 7 indicates the interval between two trains from different running directions entering a section at the same station.

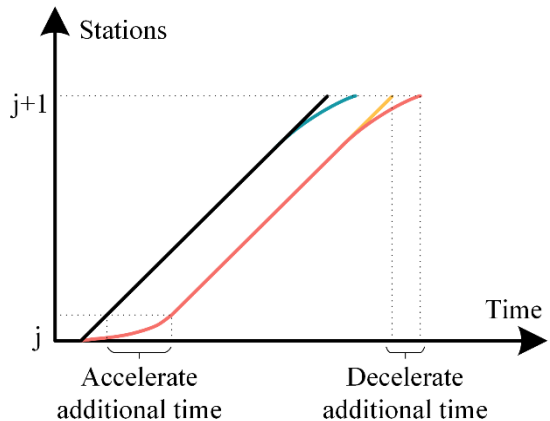

Fig.3 Four time-mileage curves of a train with different dwell settings between two stations.

When the number of trains, the first train departure and destination stations, the nominal train running time, and the first departure time of each train have been given in the railway line planning stage, if the dwell time $v_{i,j}$ of each train at each station has been determined, the intermediate variables $b_{i,j}$, $b_{i,j,t}^{station}$ and $b_{i,s(j,j(i)),t}^{section}$ can be derived from Eq. 4 and Eq. 8, as described in Eq. 11, Eq. 12 and Eq.13, respectively:

$$b_{i,j} = \begin{cases} 1, if\ v_{i,j} > 0 \\ 0, if\ v_{i,j} = 0 \end{cases} \quad (11)$$

$$b_{i,j,t}^{station} = \begin{cases} 1, if\ e_{i,j} \leq t \leq d_{i,j} \\ 0, otherwise \end{cases} \quad (12)$$

$$b_{i,s(j,j(i)),t}^{section} = \begin{cases} 1, if\ d_{i,j} \leq t \leq e_{i,j(i)} \\ 0, otherwise \end{cases} \quad (13)$$

Therefore, after each decision variable $v_{i,j}$ in the sequence $V$ is determined, we can calculate whether the sequence $V$ meets all of the constraints. If the answer is yes, $V$ determines a feasible solution, and its objective value can be calculated through the model. Otherwise, the objective value of the infeasible solution cannot be directly measured by an objective function.

3) Simulator and its dynamics model

A matrix is used as the environment and state of the simulator, and its rows and columns represent the train tracks and the discretized simulation time[56]. Here, one minute is taken as a time unit. The value of each element in the matrix can be classified into three categories, which are the $A$ and $A'$ representing no occupation by trains, the $i$ representing train occupancy, and the $B$ representing the limitation of constraints. Thus, the train operation simulation process can be transformed into the assignment process of the corresponding matrix, and the value of the matrix at any time can be reflected from the state of the simulation system at that time. The symbols about the simulator are explained in Table 2.

Table 2. Symbols defined for the simulator

| | |
|---|---|
| $row_{j,k}^{station}$ | A row vector with the length of $T_{max} + 1$. It stands for the $k$th track in station $j$, and each element corresponds to different track occupancy states at the time of index of the element. The initial value of each element in the $row_{j,k}^{station}$ is defined as a constant $A$, and $row_{j,k}^{station}(x)$ indicates the $x$th element of the vector. |
| $rows_j^{station}$ | A matrix consisting of the row vector $row_{j,k}^{station}$ for all tracks in station $j$. $rows_j^{station} = [row_{j,1}^{station}, row_{j,2}^{station}, ...]^T$, and $rows_j^{station}(x, y)$ indicates the $y$th element in the $x$th row of the vector. |
| $row_s^{section}$ | A row vector representing the track of section $s$. $row_s^{section}(x)$ indicates the $x$th element of the vector. The initial value of each element in the $row_s^{section}$ is defined as a constant $A'$. |
| $M$ | A matrix representing the tracks in the whole railway line. Its rows are staggered in the order of stations and sections in the railway line. $$M = \begin{bmatrix} rows_1^{station} \\ row_1^{section} \\ rows_2^{station} \\ row_2^{section} \\ \vdots \\ rows_{|S|}^{section} \\ row_{|J|}^{station} \end{bmatrix}_{(|J|+|S|) \times (T_{max}+1)}$$ |
| $\leftarrow$ | The operation " $\leftarrow$ " updates the value on the left with the value on the right. |
| $\otimes$ | The symbol " $\otimes$ " signifies that the simulation process fails and the simulation is finished, and the track vector is not changed. |
| $\emptyset$ | The symbol "$\emptyset$" means that the track vector is not changed. |
| $row_{j,t}^{free}$ | The operation $row_{j,t}^{free}$ means to find a row with an element with value $A$ at time $t$ (the $t+1$th column) in the matrix $rows_j^{station}$. If there exists such a row, the |

| | operation maps a row index number; and if not, the operation maps a $\otimes$. The meaning of this operation is to find the index number of the free track in the station at a certain time. |
|---|---|
| $st$ | A matrix-form state of the train operation simulator. The set of all states is recorded as $ST$, $st \in ST$. |
| $at$ | An action for the train operation simulator. Its value is same as the decision variable $v_{i,j}$. The set of all actions is recorded as $AT$. $at \in AT$. |
| $st_f$ | A terminal state indicates a failed ending of the simulator. |
| $st_s$ | A terminal state indicates a conflict-free completed ending of the simulator. |
| $st_k$ | The $k$-th state in a simulation with the simulator. |
| $at_k$ | The $k$-th action in a simulation with the simulator. |
| $tree$ | A MCTS tree consists of nodes connected with each other through edges. Nodes represent states, and edges represent actions. One edge links two nodes. The node close to the root node is called the parent node $n$, while the node far from the root node is named the child node $n'$. In addition to the corresponding state $st(n)$, a node also includes the incoming action $at(n)$, the total simulation forward $Q(n)$, and the number of accesses $N(n)$ of the node $n$. |
| $R(n)$ | The total reward obtained from the node $n$ to the end of a simulation. In particular, we record $R(n)^k$ as the reward obtained by the $k$th simulation through the node $n$. Since each state corresponds to a node, $R(st)$ correspondingly represents the total reward of the trajectory from the state $st$ to the terminal state $st_s$ or $st_f$ as well. |
| $AT(st)$ | Set of actions that can be taken in the state $st$. |
| $st(n)$ | The state corresponding to the node $n$. |
| $at(n)$ | Action to reach the node $n$. |
| $f_{sim}(st, at)$ | The subsequent state reached by executing $at$ in the state $st$. |
| $CB$ | The computing budgets. The unit of $CB$ is the number of simulations. |

In order to make object-oriented programming to realize simulation, the matrix $M$ is constructed by $rows_j^{station}$ and $rows_s^{section}$. That is, when the vector representing a track at a station or section changes, the $M$ will change as well. According to the main processes of train operation simulation, the dynamics model of the simulator can be roughly described by the simulation initialization process, the train running process, and the platforming process. The simulation processes are explained based on the dynamics equations.

According to the information obtained in the railway line planning stage, if any train $i$ starts to run on the $k$th track at the station $j$ at the time $t$, the corresponding vector $row_{j,k}^{station}$ can be updated as follows.

$$\begin{cases} row_{j,row_{j,t}^{free}}^{station}(t) \leftarrow i, & if\ row_{j,t}^{free} \neq \otimes \\ \otimes, & if\ row_{j,t}^{free} = \otimes \end{cases} \quad (14)$$

$$\begin{cases} rows_j^{station}(x,y) \leftarrow B, \forall x \in [1, C_j], \forall y \in [t-\tau_{no}, t+\tau_{no}], if\ rows_j^{station}(x,y) = A \\ \quad rows_j^{station}(x,y) \leftarrow \emptyset,\ if\ x = k,\ y = t \\ rows_j^{station}(x,y) \leftarrow \emptyset, \quad \forall x \in [1, C_j], \forall y \in [t-\tau_{no}, t+\tau_{no}], if\ rows_j^{station}(x,y) = B \\ \otimes, \forall x \in [1, C_j], \forall y \in [t-\tau_{no}, t+\tau_{no}], if\ rows_j^{station}(x,y) \neq A, B\ and\ x \neq k, y \neq t \end{cases} \quad (15)$$

According to the definition of the track vector, when the train $i$ appears on the $k$th track at the station $j$ at the time $t$, we record this element as $i$, but the following situations may occur: 1) if the time and space corresponding to this element are not occupied, the writing can be successfully done; 2) if this element has been occupied by other trains or limited by constraints, we can use $row_{j,t}^{free}$ to find other available tracks at this time of the station. If there are available tracks, the train

$i$ is written to a free track; and 3) after $row_{j,t}^{free}$ searching, if there is no available track, the initialization process fails, so does the simulation. These three cases are expressed by the three update rules in Eq. 14 respectively.

Considering the constraints described in Eq. 6, when train $i$ is written to the track vector, the station will prevent other trains from arriving at the station within $\tau_{no}$ time units. For the station track matrix $rows_j^{station}$, its constraint representation value $B$ is written into $rows_j^{station}(x,y), x \in [1, C_j], y \in [t - \tau_{no}, t + \tau_{no}]$, which is a block area of the matrix. In the writing process, as shown in Eq.15, there are several cases as follows: 1) if an element in the writing area is $A$ representing available, the value $B$ can be successfully written; 2) if the element to be written is the original $i$, skip it; 3) if the area to be written is $B$, skip it; and 4) if there are other values in the area to be written (some trains have written the values in advance), the writing of the constraint value $B$ fails, indicating that at least two trains violate the constraint as described in Eq. 6.

The successful execution of the above process for all trains indicates the simulation initialization process is also successful. To summarize the above process, it can be divided into two parts: writing train values in the available track and setting relevant constraints, as shown in Eq. 14 and Eq. 15 respectively. Fig. 4. (a) and Fig. 4. (b) show the partial schematic diagram of the matrix $M$ before and after initialization, respectively. The matrix writing method of the train movement process in the simulation process also follows this principle.

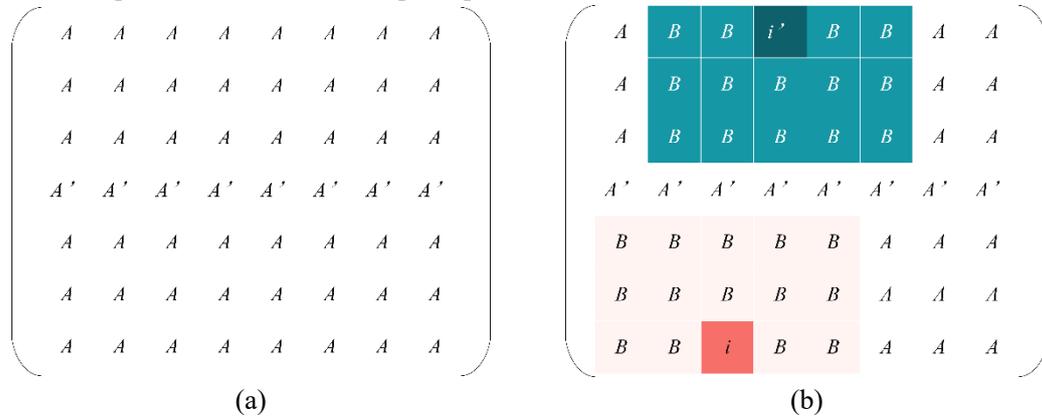

(a)     (b)

Fig. 4 Partial schematic diagram of matrix $M$. (a) is the matrix before initialization, and (b) is the matrix after initialization. Train numbers $i$ and $i'$ are set in the process described in Eq. 14, and the surrounding area with the value of $B$ is set in the constraint writing process described in Eq. 15.

There are three steps in the train running process, which are departing from the previous station, section running, and arriving at the next station[56]. The dynamics equations are demonstrated as the Eq. 16, Eq. 17 and Eq. 18, respectively. It is set that the train $i$ departs from the station $j$ on the $k$th track at time $t^d$ after it dwells $v_{i,j}$ minutes from arriving at this station at the time $t$, enters the section $s = s(j, j(i))$, and then arrives at the station $j(i)$ ahead at the time $t^e$. According to Eq. 4 and Eq.8, $t^d = v_{i,j} + t$, $t^e \equiv t^d + t_{i,s(j,j(i))}^{nom} + b_{i,j} t_{i,j}^{acc} + b_{i,j(i)} t_{i,j(i)}^{dec}$. The corresponding vector $rows_j^{station}$, $row_s^{section}$, and $rows_j^{station}$ are updated as below.

$$\begin{cases} row_{j,k}^{station}(x) \leftarrow i, \forall x \in (t, t^d], if\ row_{j,k}^{station}(x) = A\ or\ B \\ row_{j,row_{j,x}^{free}}^{station}(x) \leftarrow i, \forall x \in (t, t^d], if\ row_{j,k}^{station}(x) \neq A\ or\ B, row_{j,x}^{free} \neq \otimes \\ \otimes, \forall x \in (t, t^d], if\ row_{j,k}^{station}(x) \neq A\ or\ B, row_{j,x}^{free} = \otimes \end{cases} \quad (16)$$

When a train dwells at a place, the train number is written by the dynamics equation Eq.16 to the subsequent elements of the vector according to its dwell time, and the existing values $A$ and $B$ will not affect the writing process. However, if another train number already exists, it will look for an available track vector in the station at that time. If an available track vector exists, the writing process continues; and if not, then it means that the station has no available track at that time and the simulation fails.

$$\begin{cases} row_s^{section}(x) \leftarrow B, \forall x \in [t^d - \tau_d, t^d), if\ row_s^{section}(x) = A' \\ row_s^{section}(x) \leftarrow \emptyset, \forall x \in [t^d - \tau_d, t^d), if\ row_s^{section}(x) = B \\ \otimes, \forall x \in [t^d - \tau_d, t^d), if\ row_s^{section}(x) \neq A'\ or\ B \end{cases} \quad (17)$$

For the constraints described in Eq. 7, when a train departs from a station and enters the front section, another train is not allowed to enter the station within the $\tau_d$ time range. Therefore, we write the constraint value $B$ to the $\tau_d$ time range in this area. As shown in Eq. 17, if an element with value $A'$ or $B$ already exists at the writing location, the writing process will not be affected; and if a train number already exists, the writing fails, indicating that the constraint of Eq. 7 has been violated.

$$\begin{cases} row_s^{section}(x) \leftarrow i, \forall x \in [t^d, t^e], if\ row_s^{section}(x) = A' \\ \otimes, \forall x \in [t^d, t^e], if\ row_s^{section}(x) \neq A' \end{cases} \quad (18)$$

When a train departs from a station and enters a section, the running time of the train is determined according to Eq. 8. As shown in Eq. 18, we write train numbers to the section vector, and if its subsequent elements are not $A'$, the writing process fails, which means that the train running in the section does not meet the constraint described in Eq. 7 or Eq. 10.

It is also necessary to think about whether to add additional time for train acceleration and deceleration in the course of calculating the running time of a train in a section. In this paper, all unexecuted actions are obtained randomly, as shown in Sec. III, so the method of first randomly determining the undetermined actions in the whole action sequence and then executing them one by one is utilized, and whether the additional time for acceleration and deceleration needs to be calculated in the simulation can be determined by the known subsequent actions, as shown in Eq. 11.

$$\begin{cases} row_{j(i),row_{j(i),t^e}^{free}}^{station}(t^e) \leftarrow i,\ if\ row_{j(i),t^e}^{free} \neq \otimes \\ \otimes, if\ row_{j(i),t^e}^{free} = \otimes \end{cases} \quad (19)$$

$$\begin{cases} rows_{j(i)}^{station}(x,y) \leftarrow B, \forall x \in [1, C_{j(i)}], \forall y \in [t^e - \tau_{no}, t^e + \tau_{no}], if\ rows_{j(i)}^{station}(x,y) = A \\ rows_{j(i)}^{station}(x,y) \leftarrow \emptyset,\ if\ rows_{j(i)}^{station}(x,y) \neq A \end{cases} \quad (20)$$

When the train arrives at the front station, the operation of the station matrix is similar to that in the initialization process, as presented in Eq. 19 and Eq. 20. The train running process in the train operation simulation is shown in Fig. 5.

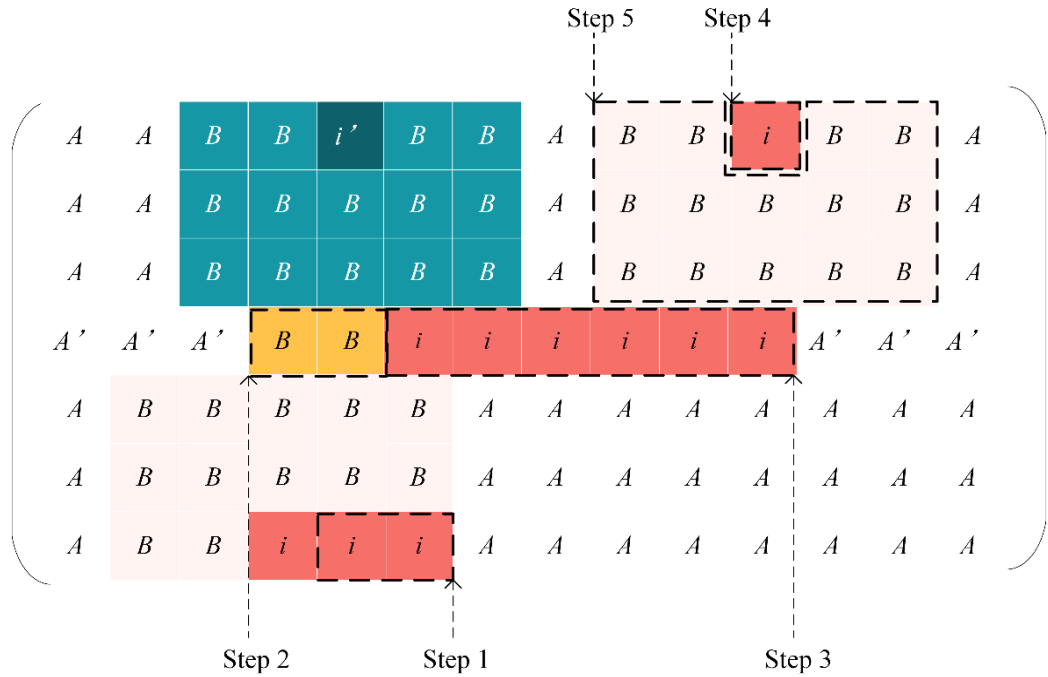

Fig.5 Schematic diagram of a train running process in the matrix $M$. Step 1 refers to the process of train dwelling, as described in Eq. 16; Step 2 is to set section constraints, as indicated in Eq. 17; Step 3 is the section running process of the train, as demonstrated in Eq. 18; and Step 4 means the train arrives at the station in advance, as shown in Eq. 19. Then, constraints on the surrounding areas are set after the train arrives at the station in advance, as described in Eq. 20.

Another process of train operation simulation is the platforming process. It is assumed that the tracks at all stations are in an equal status, and as explained in Eq. 9, the track capacity in all stations has been guaranteed, so the platforming process in the simulation is to arrange each train number in each station matrix $rows_j^{station}$ in the matrix $M$ into a unified track vector in order to maintain the continuity of track allocation for each train (Fig. 6). This process is easy to be implemented after the train running process is completed, so detailed explanations will not be given here.

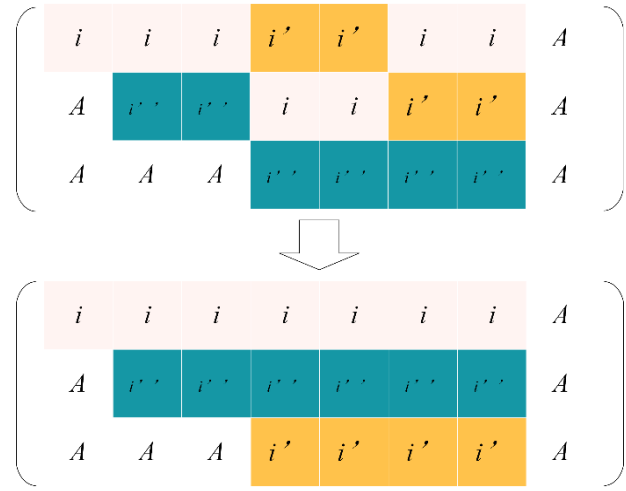

Fig.6 Partial schematic diagram of a station matrix before (upper) and after (lower) the platforming process.

As a matter of fact, the dynamics equations of the train running process are to record the train trajectories and constraints in the matrix $M$ where the horizontal and vertical axes represent the time and the track respectively. The writing process is also the constraint checking process, which is important for the simulation. When a train arrives at the destination station from its first departure station, the writing and checking process continues for the next train until the simulation is completed or fails. Once the simulation process completed, we can easily derive the train timetable and station track allocation plan according to $M$. The flow diagram of the whole simulation process is shown in Fig. 7.

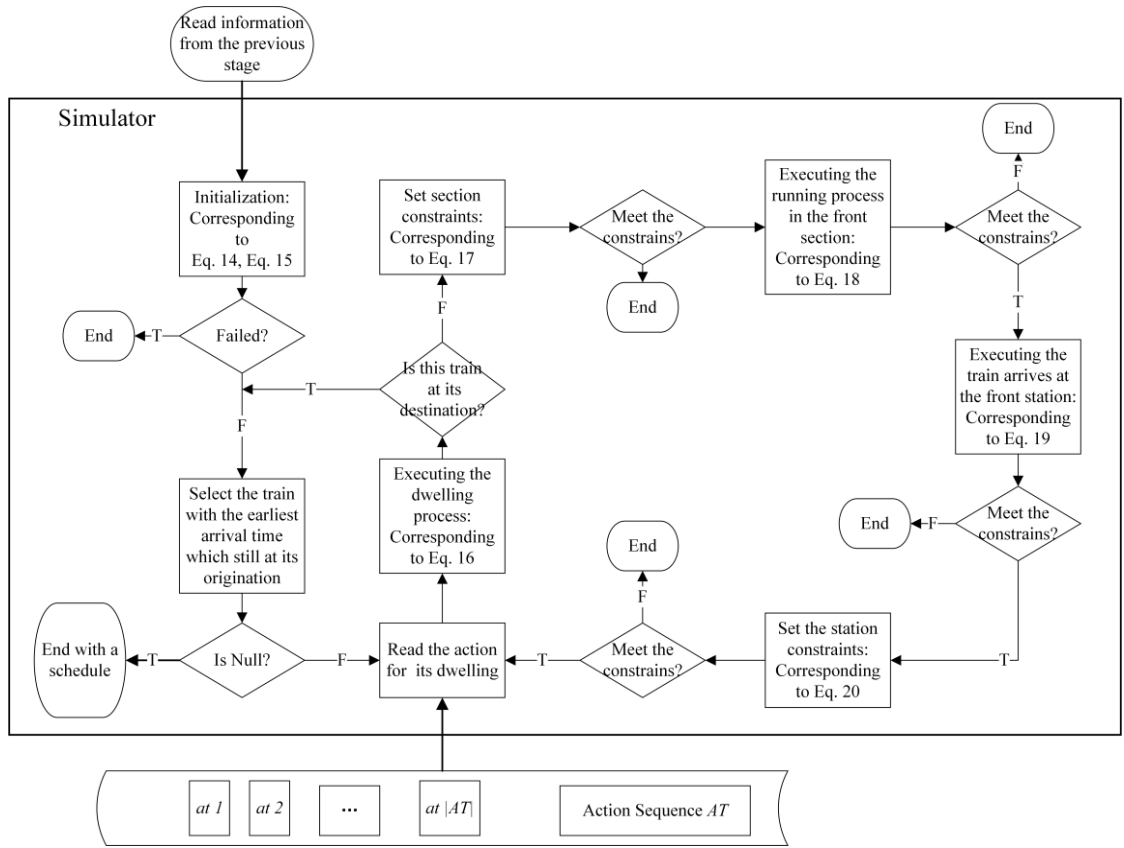

Fig. 7 Flow diagram of the whole simulation process

The matrix $M$ at each simulation time is regarded as the state $st$ of the simulator, and all possible states are recorded as the state set $ST$. In particular, the states of the failed simulation and of simulation completed are recorded as terminal states $st_f$ and $st_s$ respectively. The first state after the initialization process is recorded as the initial state $st_0$, and the subsequent states are subsequently recorded as $st_1, st_2, \ldots, st_f/st_s$. We take the dwell time of each train as an action $at$, and all actions are recorded as the action set $AT$. Likewise, all actions in a simulation are recorded as $[at_0, at_1, ...]$.

**B The Vanilla MCTS and UCT**

Realized by building a multi branch tree, the core of MCTS is to evaluate the potential subsequent state before each action so as to select the action that can achieve the best subsequent state. The MCTS framework generally consists of four phases, *Selection*, *Expansion*, *Simulation* and *Back Up*, as shown in Fig. 8.

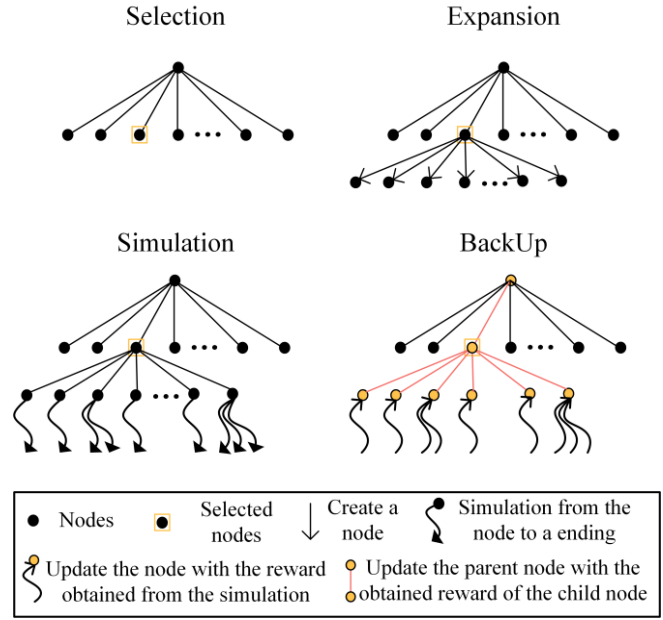

Fig. 8 Four phases of the MCTS framework

1) *Selection*: Starting at the root node, a child node selection policy is recursively applied to descend through the tree until the most urgent expandable node is reached. A node is expandable if it represents a nonterminal state and has unvisited children[13].
2) *Expansion*: One child node is added to expand the tree, according to the available actions[13].
3) *Simulation*: A simulation is run from the new node according to the default policy to produce an outcome[13].
4) *Back Up*: The simulation result is backpropagated through the selected nodes to update their statistics[13].

In order to evaluate each node, the $Q$ value is defined as the mean outcome of all trajectories from the node to potential leaf nodes. The $Q$ value of each node is described in Eq. 21 according to the UCT algorithm,

$$Q(n) \stackrel{\text{def}}{=} \frac{\sum_{k=1,2,...,CB_n} R(n)^k}{k} \quad (21)$$

where $CB_n$ is the number of the simulations through the node $n$, and $CB_n \in \mathbb{N}, 0 < CB_n \leq CB$. Since the computing time of MCTS algorithms depends on the time and number of simulation execution, we use the computing budget $CB$ to describe the computing resources (computing time and parallel simulation threads) given to the algorithm. When the state space is large, it is unlikely to build a tree containing all states. The MCTS only expands one of the most promising branches through the *Selection* and *Expansion* phases, and selecting one of the reachable child nodes is done through the *Simulation* and *Back Up* phases. Theoretically, accurate evaluation of the best node requires infinite simulations, but pratically the computing budget is always limited. The UCT algorithm gives an effective computing budget allocation strategy, which enables better computing performance within the limited computing budget. In light of its good scalability and computing performance, the UCT is a typical algorithm of MCTS framework. The general MCTS framework and the UCT algorithm are shown respectively as follows.

Algorithm 1. The General MCTS Framework

| **Function** $MCTS\ FRAMEWORK(st)$ |
|---|

  Create node $n$ with state $st$
  **While** the state of $n$ is not $st_f$ or $st_s$
    **While** within $CB$:
      $n_{tmp} \leftarrow TreePolicy(n)$ //Implemented the *Selection* and *Expansion* process
      $R(n_{tmp}) \leftarrow DefaultPolicy(st(n_{tmp}))$ //Implemented the *Simulation* process
      $BackUp(n_{tmp}, R(n_{tmp}))$ //Implemented the *Back Up* process
    **End While**
    $at \leftarrow at(BestChild(n, 0))$ (22)
    $n \leftarrow BestChild(n, 0)$
  **End While**
**End**

Algorithm 2. The UCT Algorithm

**Function** $UCT$ //Same with the $MCTS\ FRAMEWORK(st)$
**Function** $TreePolicy(n)$
  **While** $st(n) \neq st_f\ or\ st_s$
    **If** $n$ not fully expanded
      Return $Expand(n)$
    **Else**
      $n \leftarrow BestChild(n, Cp)$
   Return $n$
**End**
**Function** $Expand(n)$
  Choose $at \in untried\ actions\ from\ AT(st(n))$
  Add a new child node $n'$ to $n$
    With $st(n') = f_{sim}(st(n), at)$
    And $at(n') = at$
  Return $n'$
**End**
**Function** $BestChild(n, Cp)$
  **Return** $argmax_{n' \in children\ of\ n} UCB(n', Cp), UCB(n', Cp) = (\frac{Q(n')}{N(n')} + Cp\sqrt{\frac{2lnN(n)}{N(n')}})$ (23)
**End**
**Function** $DefaultPolicy(st)$
  **While** $st$ is nonterminal state
    Choose $at \in AT(st)$ uniformly at random //Choose next random action from the predetermined action sequence.
    $st \leftarrow f_{sim}(st, at)$
  **End**
  Return $R(st)$
**End**
**Function** $BackUp(n, R)$
  **While** $n$ is not null
    $N(n) \leftarrow N(n) + 1$
    $Q(n) \leftarrow Q(n) + R$
    $n \leftarrow parent\ node\ of\ n$
  **End**
**End**

  For the MCTS framework, it can be seen from Algorithm 1 that the $TreePolicy$ is adopted to construct a tree and select the nodes to be evaluated, and the $DefaultPolicy$ is used to evaluate a node. After evaluation, when the optimal action has been selected by Eq. 22 in the current state, the current node is replaced with the optimal child node after the optimal action is executed, and this process is implemented recursively until the leaf node is reached. As shown in Eq. 23, the UCT resolves the Exploration-Exploitation dilemma by balancing the number of visits to a node with the $Q$ value of the node. The node selection and evaluation method of MCTS makes the selection of overall optimal actions present a "Best-first" strategy. Through a fixed number of simulations, other unselected actions will be discarded after the optimal actions are selected, which makes it easy to evaluate between the computing budget and the computational accuracy.

According to the $DefaultPolicy$ function, the actions in a simulation process are always randomly determined. Therefore, we use the predetermined method described previously to practically construct the entire unexecuted random actions instead of randomly choosing one at a time. That is to say, before implementing a simulation, we first randomly generate an action sequence $[at_0, at_1, ..., at_{|V-1|}]$ based on the length of the action sequence $V$ we need to make decisions, and then this action sequence is executed sequentially during the simulation process until all actions in the action sequence are executed or the simulation termination state is reached. This covers the shortage of the common RL method which cannot predetermine the subsequent actions to decide whether a train needs to consider the additional time for its acceleration and deceleration in a simulation.

## III. Methodology

In this section, we first introduce how the UCT in a planner solves the TTP of single-track railways and how to make the UCT better applicable to the TTP, and propose the improved solving algorithms UCT_MAX1 and UCT_MAX2. We also put forward that how to use the heuristic strategy CSAV to enhance the searching efficiency of the basic algorithms in a planner when the computing budget is insufficient. Then, we describe the settings of VAF in a learner, and explain the collection and preprocessing methods of training datasets, as well as how VAF learns training data. Finally, the integrated method UCT_VAF combining a planner and a learner is described.

### A. The Planner: Using MCTS algorithms to solve the TTP instances

The UCT, a popular MCTS algorithm, is adopted as the basic algorithm for solving the TTP in a planner to explain how the planner works. Based on the characteristics of TTP, we propose two improved algorithms UCT_MAX1 and UCT_MAX2. Theoretically, the optimal solution to the TTP can be obtained through the three algorithms when the computing budget is sufficient. In actual scenarios, when the computing budget is insufficient (or the size of problem instances is large), how to use a heuristic strategy to combine with the above three algorithms to improve the solving efficiency is what we are going to explain next.

**UCT for the TTP**

Since train timetables can be determined by deciding each element in the decision sequence $V$, the objective function such as Eq. 1 can be optimized by determining the decision sequence $V$. Correspondingly, we regard each variable $v_{i,j}$ in decision sequence $V$ as an action $at$, and then we can optimize the objective function by determining an optimal action sequence through the MCTS.

When the initial information described by a single-track railway TTP is initialized by a simulator, its initial state $st_0$ is used to build the root node $n_0$ of a tree. According to Eq. 5, each node may have $V_{max}$ child nodes, while a tree has $|V|$ layers of child nodes. We construct the corresponding state $st_{k+1}$ of each child node according to different action inputs $at$ in the simulation with $st_{k+1} = f_{sim}(st_k, at)$. In addition to the incoming actions $at(n)$ and $st(n)$, each node $n$ also sets the access number variable $N(n)$ and value variable $Q(n)$. The constructed MCTS tree is shown in Fig. 9, where $f_{sim}(st, at)$ represents a simulation step of the simulator, $st$ is the state of the simulator, and $at \in [0, V_{max}]$ denotes the input action. The corresponding output of this simulation step is the subsequent state reached.

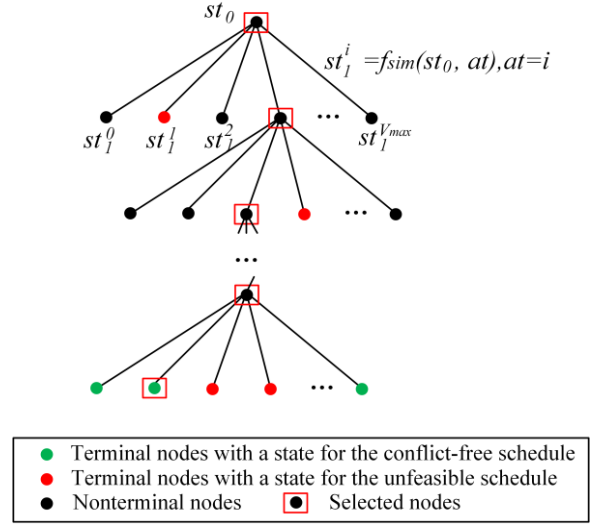

Fig. 9 A diagram of a MCTS tree for TTP of single-track railways.

If the state of a node is $st_s$ or $st_f$, it means that the simulation process gets a conflict-free or unfeasible timetable accordingly. For a conflict-free timetable, we can get the corresponding total train travel time $T_{travel}$ based on the statistics of the trajectory of each train. If a timetable conflict occurs, its corresponding objective value is defined as 0, otherwise the reward value $R(st_s)$ for each state $st_s$ is defined as described in Eq. 24.

$$R(st_s) = \frac{T'_{travel} - T_{travel}}{T'_{travel}} \quad (24)$$

$T'_{travel}$ is the possible maximal total travel time for the problem instance, and $T'_{travel} < \infty$ and $R(st_s) \in [0,1]$ with the limited dwell time. This normalized reward setting makes the states of different instances comparable and the algorithm more applicable.

The optimal action sequence is selected by Eq. 22. Once the computing of the algorithm completed, the solution of the problem instance is determined. In addition, with the sufficient computing budget $CB$ of the algorithm, all paths from each selected node to all leaf nodes are estimated, which is achieved through randomness in the $DefaultPolicy$. Therefore, the asymptotic optimality of the algorithm is guaranteed.

Taking a multiway search tree into consideration, if a Full multiway search tree is constructed, there are $1 + (V_{max} + 1)((V_{max} + 1)^{|V|} - 1)/V_{max} \approx V_{max}^{|V|}$ nodes in the tree. The Full multiway search tree needs $V_{max}^{|V|}$ simulations to evaluate each node so as to select the optimal action sequence, while the UCT algorithm only requires $CB * |V|$ simulations, and the corresponding computing time is able to be greatly reduced.

**Improving UCT with the Max Operator**

As shown in Sec. III-A, the TTP of single-track railways is solved with the UCT. However, in the case of the limited computing budget, the UCT needs to be further improved to be applied in the TTP. First, from Algorithm 2, the function $DefaultPolicy$ evaluates each node by generating random actions that obey a uniform distribution. Assuming that $CB_n$ action sequences $AT_{default} = (\bigcup_{z=1,2,\dots,CB_n}[at_0, at_1, \dots, at_k]^z)$ are performed to estimate the node value of $n$, the probability that each action in the sequences is assigned any specific action value is $1/|AT|$, the reached terminal state of these actions is $st^z$, and then the estimated value $Q(n)$ of this node is shown in Eq. 25.

$$Q(n) \stackrel{\text{def}}{=} \frac{\sum_{k=1,2,\dots,CB_n} R(n)^k}{k}$$
$$\cong \frac{\sum_{z=1,2,\dots,CB_n}(\mathcal{P}(st^z \text{ is } st_s)R(st^z) + \mathcal{P}(st^z \text{ is } st_f)0)}{CB_n}$$
$$= \frac{\sum_{z=1,2,\dots,CB_n} \mathbb{E}(R(st^z)|st^z \text{ is } st_s)}{CB_n} \quad (25)$$
$$= \underset{CB_n, at \sim U, st^z \in st_s}{mean}\mathbb{E}(R(st^z))$$

$$= \frac{\underset{CB_n, at \sim U, st^z \in st_s}{mean}\mathbb{E}(T_{travel})}{-T'_{travel}}$$

$\mathcal{P}(st^z \text{ is } st_s)$ and $\mathcal{P}(st^z \text{ is } st_f)$ denote the probability that the terminal state $st^z$ is a simulation success and a simulation failure, respectively, and $at \sim U$ illustrates that the actions in these simulations are randomly selected from $AT$ with $1/|AT|$ probability. $st^z \in st_s$ marks the simulations ended with a successful state. When $CB_n$ is large enough, the operator $'\cong'$ in Eq. 25 could take the $'='$ operator according to the Theorem of Large Numbers. In the case of simulations with successful terminal states, the $T_{travel}$ increases monotonously with the sum of elements of an action sequence $[at_0, at_1, \ldots, at_k]^z$, so Eq. 26 can be derived.

$$Q(n) \propto -\underset{CB_n, at \sim U, st^z \in st_s}{mean}\mathbb{E}\left(\sum [at_0, at_1, \ldots, at_k]^z\right) \quad (26)$$

According to the Central Limit Theorem, the sum of random action sequences follows a normal distribution with the mean value of the action sequence approximately, so the UCT algorithm applied in the TTP tends to use a dwell time action sequence with an average value of all actions to evaluate nodes of a tree, while combinations of actions with a smaller dwell time are less likely to be chosen in the *Simulation* process. Unlike the game of Go[48], our goal is to minimize the total travel time of all trains, and we would like to explore more combinations of actions that represent a potential smaller total train dwell time.

To achieve this goal, the Max operator is used to improve the $DefaultPolicy$ and evaluation function UCB, as shown in Algorithm 3 and Algorithm 4. The improved UCT_MAX1 algorithm is same as the UCT algorithm applied to TTP, but the function $MCTS\ FRAMEWORK$ and $BackUp$ are modified.

Algorithm 3 UCT_MAX1

---

**Function** $UCT\_MAX1\ (st)$
    Create node $n$ with state $st$ with a new field $Q_{Max}(n) = 0$
    **While** the state of $n$ is not $st_f$ or $st_s$
        **While** within $CB$:
            $n_{tmp} \leftarrow TreePolicy(n)$
            $R(n_{tmp}) \leftarrow DefaultPolicy(st(n_{tmp}))$
            $BackUp(n_{tmp}, R(n_{tmp}))$
        **End While**
        $at \leftarrow at\left(argmax_{n' \in children\ of\ n} Q_{Max}(n')\right)$
        $n \leftarrow argmax_{n' \in children\ of\ n} Q_{Max}(n')$
    **End While**
**End**
**Function** $TreePolicy(n)$ //Same with it in $UCT$
**Function** $Expand(n)$ //Same with it in $UCT$
**Function** $BestChild(n, Cp)$ //Same with it in $UCT$
**Function** $DefaultPolicy(st)$ //Same with it in $UCT$
**Function** $BackUp(n, R)$
    **While** $n$ is not null
        $N(n) \leftarrow N(n) + 1$
        $Q(n) \leftarrow Q(n) + R$
        $Q_{Max}(n) \leftarrow Max(Q_{Max}(n), R)$ //A new variable for nodes
        $n \leftarrow parent\ node\ of\ n$
    **End**
**End**

---

UCT_MAX1 still uses the $UCB$ function in UCT to resolve the Exploration and Exploitation dilemma, but it also records the best performance of each node in the simulation and selects the node that performs best in the evaluation when choosing the next action. Although UCT_ MAX1 does not change the distribution of action sequences in the *Simulation* process, it does not directly use $Q(n)$ to make decisions in the actual action selection, thereby weakening the impact of the normal distribution of the sum of action sequences on the algorithm performance.

Furthermore, in order to make the child nodes with smaller dwell time represented in the *Simulation* process get more explorations, UCT_MAX2 is proposed based on the UCT_MAX1 algorithm through modifying the evaluation function $UCB$, as shown in Algorithm 4.

Algorithm 4 UCT_MAX2
---
**Function** $UCT\_MAX2\ (st)$ //Same with the $UCT\_MAX1(st)$
**Function** $TreePolicy(n)$ //Same with it in $UCT\_MAX1$
**Function** $Expand(n)$ //Same with it in $UCT\_MAX1$
**Function** $BestChild(n, Cp)$

$$UCB_{MAX}(n', Cp) = Q_{Max}(n') + Cp\sqrt{\frac{2lnN(n)}{N(n')}} \quad (27)$$

**Return** $argmax_{n' \in children\ of\ n}\ UCB\_MAX(n', Cp)$
**End**
**Function** $DefaultPolicy(st)$ //Same with it in $UCT\_MAX1$
**Function** $BackUp(n, R)$ //Same with it in $UCT\_MAX1$
---

It can be seen from Eq. 27 that the UCT_MAX2 uses $Q_{Max}$ to participate in the computing budget allocation on each child node, allowing nodes with the best performance in the *Simulations* process to get more simulations than the nodes with the best average performance. When the computing budget is enough, UCT_MAX1 and UCT_MAX2 algorithms degenerate into dynamic programming algorithm, and their asymptotic optimality still exists.

**Improving Searching efficiency with a Heuristic Strategy CSAV**

By observing a state matrix $M$ that represents a conflict-free simulation for 52 trains on a track with ten stations in about 1600 minutes, it is found that about 20% of its elements are occupied by various constraints and train trajectories. Here is a simple deduction: assuming that each element in an action sequence randomly chooses a dwell time, the probability of a conflict occurring is 10% after the action is executed, and the probability that a sequence with about 520 random actions can successfully execute a conflict-free running simulation is $0.9^{520} \approx 10^{-24}$, which means that we need to simulate about $10^{16}$ times to have a 90% probability of completing a conflict-free simulation. It is unacceptable both in computing time and space.

When Algorithms 2 to 4 are applied to a large size of the TTP instances, all child nodes that are close to the root node will fail to reach a leaf node with a conflict-free terminal state within the computing budget in the *Simulation* process. The failure indicates that the $Q(n)$ of all child nodes are zero, causing that the MCTS algorithm fails in actual scenarios (theoretically, this situation will not happen with the sufficient computing budget).

Although it is difficult to randomly choose an action sequence to simulate a conflict-free train timetable directly, we can still evaluate a conflicting train timetable. On one hand, the action sequence that can reach a deeper layer without conflicts is better than that can only reach a shallow layer. On the other hand, for these action sequences that can reach the same layer, we think the action sequences with less total train travel time are better. In this regard, we present a method of counting the steps of the actions that can be executed without conflicts as a main evaluation factor, and the total train travel time corresponding to the action sequence that has been executed without conflicts is considered as a secondary factor. Next, a weighted sum operation is performed, whose value is used as a measure to evaluate an action sequence based on the general UCT algorithms. The proposed algorithm is shown in Algorithm 5.

The UCT_CSAV adopts a similar MCTS framework as the general UCT algorithms, but a variable $cs$ is added to indicate the step number of the executed conflict-free actions in its *Simulation* process, as shown in Eq. 28. A variable $R^{CSAV}(st)$ in the form of weighted rewards is also added to demonstrate the reward obtained by a simulation with a terminal state $st$, as shown in Eq. 29, where the variable $T\_travel$ represents the executed total conflict-free train travel time until a conflict occurs, the parameters $\alpha$ and $\beta$ denote the weight of $cs$ and $\overline{R}(st)$ respectively, and $\beta$ is much smaller than $\alpha$. Similar to $R(st)$, the $\overline{R}(st)$ is a generalized reward function.

It can be seen from Eq. 30 that when a random action sequence fails to reach the leaf node containing a state $st_s$ in *Simulation* process, $\overline{R}(st)$ would still evaluate the action sequence according to this heuristic strategy and give the sequence a value that only contains the sum of $\alpha \cdot cs$ and $\beta \cdot \overline{R}(st)$. Since $\beta$ is far smaller than $\alpha$, $\beta \cdot \overline{R}(st)$ is meaningful only when the $cs$ of action sequences are the same. Additionally, when the feasible solutions cannot be found, the heuristic strategy focuses on guiding the algorithm to select actions that can potentially complete more conflict-free steps. On the contrary, when the leaf nodes with a feasible solution can be found, the item $\alpha \cdot cs$ is invalid (all the action sequences that can find a feasible solution have the same value of $cs$), $\beta \cdot \overline{R}(st)$ has the same meaning as $R(st)$, and the performance of the algorithm is similar to the general UCT algorithms. Hence, when the computing budget of UCT_CSAV is sufficient (enough feasible solutions can be found), the algorithm has the same asymptotic optimality as the general UCT algorithms.

---
Algorithm 5 UCT_CSAV
---

**Function** $UCT\_CSAV\ (st)$
    Create node $n$ with state $st$
    **While** the state of $n$ is not $st_f$ or $st_s$
        **While** within $CB$:
            $n_{tmp} \leftarrow TreePolicy(n)$
            $R^{CSAV}(n_{tmp}) \leftarrow CSAVPolicy(st(n_{tmp}))$
            $BackUp(n_{tmp}, R^{CSAV}(n_{tmp}))$
        **End While**
        $n \leftarrow BestChild(n, 0)$
    **End While**
**End**
**Function** $TreePolicy(n)$ //Same with it in general UCT algorithms
**Function** $Expand(n)$ // Same with it in general UCT algorithms
**Function** $BestChild(n, Cp)$ // Same with it in general UCT algorithms
**Function** $CSAVPolicy(st)$
    Initial $cs = 0$ //A new variable
    Initial $T\_travel = 0$ //A new variable
    **While** $st$ is nonterminal state
        Choose $at \in AT(st)$ uniformly at random
        $st \leftarrow f_{sim}(st, at)$
        **If** $st$ is nonterminal state
            $cs \leftarrow cs + 1$ (28)
            $T_{travel} \leftarrow Statistics\ of\ total\ train\ travel\ time\ T_{travel}\ executed$
        **End**
    **End**
    $\overline{R}(st) = \dfrac{T'_{travel} - T\_travel}{T'_{travel}}$ (29)
    $R^{CSAV}(st) = (1 - \alpha - \beta)R(st) + \alpha \cdot cs + \beta \cdot \overline{R}(st)$ (30)
    **Return** $R^{CSAV}(st)$
**End**
**Function** $BackUp(n, R^{CSAV})$ // Same with it in general UCT algorithms

---

## B. The Learner: Learning knowledge obtained by the planner

The purpose of the planner is to solve TTP instances, but it does not have the ability to learn. In this section, we are going to describe the network setting of the learner, the process of data collection and preprocessing, and the network training process. Besides, the integrated method of combining a learner and a planner is described in detail.

**Learning Knowledge from MCTS with CNN**

It can be known from algorithms 1 to 5 that the value of each node is estimated by the values of the subsequent nodes that could be reached, but essentially, this sampling method is only a statistical means, the core of which determines the value of the node is the value of the state itself. Because estimating the value of a state through a large number of simulations is computationally intensive, in this paper, a supervised learning method is presented to learn and generalize the states that have been estimated by the MCTS method, and a non-linear function is obtained to quickly evaluate similar states. We use the deep convolutional network, a nonlinear function, as the VAF, and its network structure is shown in Figure 10.

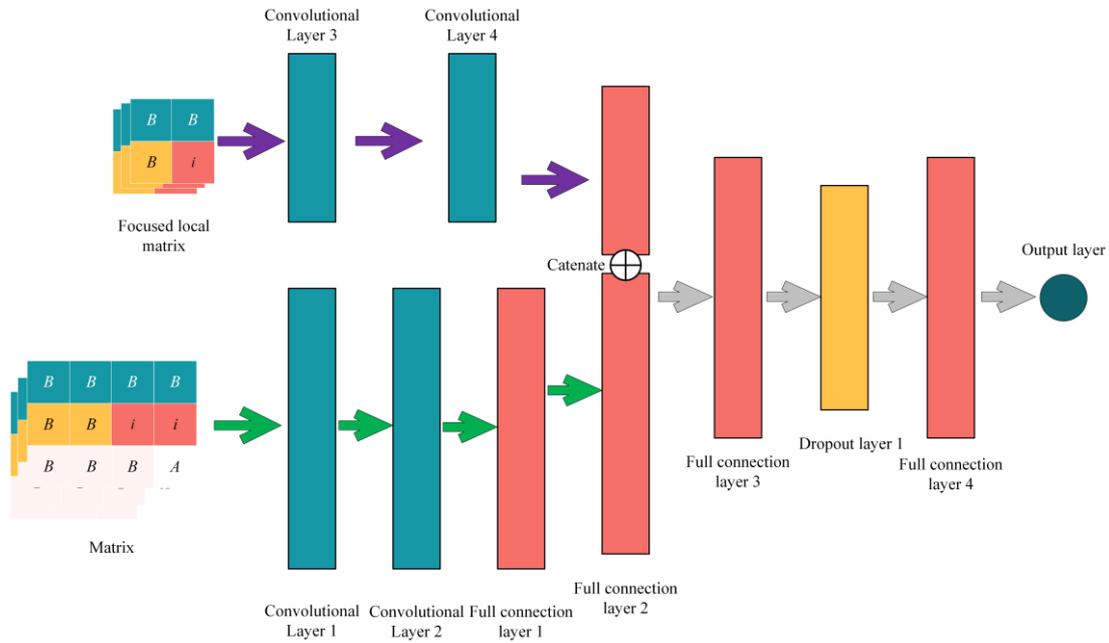

Fig. 10 Structure Diagram of the CNN of the VAF.

As indicated by Fig. 10, the CNN has two data input channels, which corresponds to the input of the entire matrix data and the focused local matrix, and each of them has three two-dimensions matrixes. Since the convolutional layer has a proportional relationship between the accuracy of data feature extraction and the size of the generated feature maps, VAF uses $3 \times 3$ convolutional kernels with higher accuracy for the focused local matrix, while $5 \times 5$ convolutional kernels with lower accuracy and larger stride are used for the entire matrix data, thus balancing the computational costs and the computational accuracy. After the two data processing paths are processed separately, VAF uses a full connection layer to connect the two data processing paths together, and a Dropout layer[57] is used in the subsequent full connection layers to make VAF more robust. The output of VAF is the value of a state, so the output layer of the proposed deep neural network is a single neuron and outputs a scalar.

As mentioned above, the input data of VAF is the state of some MCTS tree nodes, and the value corresponding to each state for supervised training is also generated by the MCTS framework, as shown in Fig. 11.

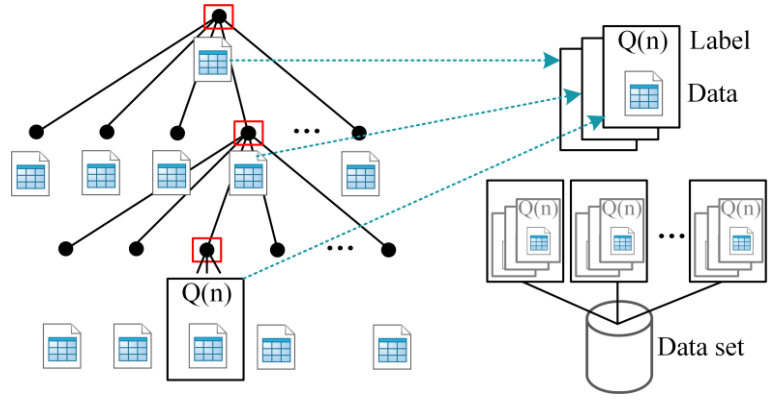

Fig. 11 Diagram of using MCTS to generate state data and corresponding values.

The *Simulation* phase for a node completed, its child nodes have been fully evaluated to some degree, so we only collect the state matrixes of the child nodes of the parent node selected through the *Selection* phase. The state matrixes are taken as the data for supervised training, and the $Q$ values of the corresponding nodes are used as labels. Since trains have different running speeds, the date preprocessing method in AlphaGo is taken as a reference, and the method of stacking the current state, the state of the parent node, and the state of the grandparent node is adopted to acquire the history of the state matrixes in the simulation. Then, the acquired matrix data and label data are stored in a dataset to support the training and testing of supervised learning for VAF.

In view of the large size of the entire matrix data, if CNN is being directly used, it would consume a lot of computing resources to extract its features, and would be difficult to identify useful features from different interference. Therefore, we separately extract the spatiotemporal area around the train being decided represented by the current state in the simulator as an input of VAF. The extracted area is displayed in Fig. 12, and the extracting window moves with the position of the train being decided represented by each state.

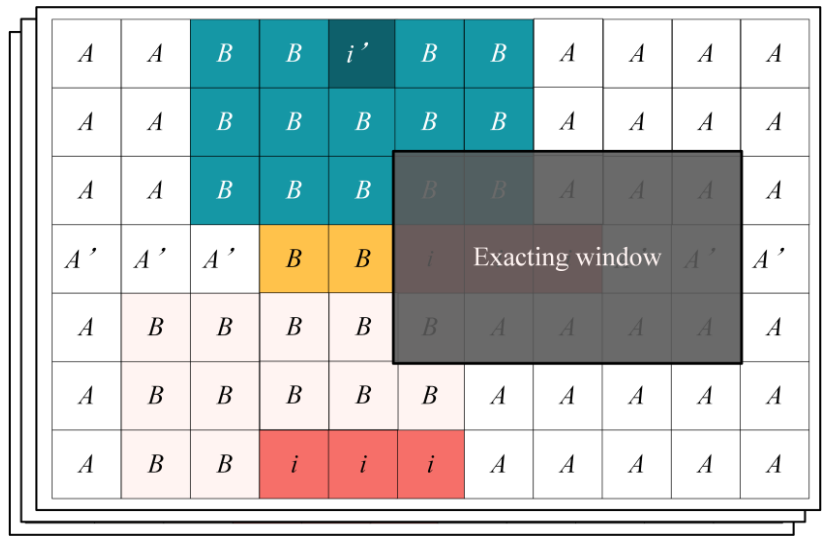

Fig. 12 The schematic diagram of the extracting window for the focused local matrix.

The VAF establishes a mapping between the state and its value, which is recorded as $f_{vaf}(st) \to Q^{vaf}(st)$ or $f_{vaf}(n) \to Q^{vaf}(n)$, representing the value estimation $Q^{vaf}$ for a state $st$ or the value $Q^{vaf}$ for the state in a node $n$.

The MCTS provides not only a training dataset for the supervised learning, but also a credible label set for it. Besides, the deep neural network is good at extracting the features of these data and summarizing a certain degree of commonness among different data. The more sufficient datasets and training time, the more powerful and applicable VAF can be obtained, which means that the existing computing resources can be used efficiently.

## C The Integrated Framework

Since the training data of $f_{vaf}$ is obtained from MCTS algorithms, the most natural application method is to embed it into the MCTS to form a Deep MCTS method, and, without doubt, there are many other potential uses of this function. In this paper, we use the Warm up method to pre-evaluate the $Q$ values of the child nodes in the *Expansion* phase with $f_{vaf}$ in advance. The pre-evaluation is able to reduce the computational cost in the *Simulation* phase and improve the accuracy of the computing budget allocation for potential good nodes. There are multiple feasible algorithms in a planner that can be integrated with a learner, so the method of combining a learner with a planner is called the integrated framework. The integrated algorithm UCT_VAF with applying VAF to the general UCT algorithms is shown in Algorithm 6.

Algorithm 6. The UCT_VAF Algorithm
***
**Function** $UCT\_VAF(st)$ // Same with it in general UCT algorithms
**Function** $TreePolicy(n)$// Same with it in general UCT algorithms
**Function** $Expand(n)$//Same with it in $UCT$
    Choose $at \in untried\ actions\ from\ AT(st(n))$
    Add a new child node $n'$ to $n$
        With $st(n') = f_{sim}(st(n), at)$
        And $at(n') = at$
    $Q(n') \leftarrow f_{vaf}(st(n'))$ //Warm up the new child node
    Return $n'$
**End**
**Function** $BestChild(n, Cp)$ //Same with it in general UCT algorithms
**Function** $DefaultPolicy(st)$ //Same with it in general UCT algorithms
**Function** $BackUp(n, R)$ //Same with it in general UCT algorithms
***

Although in theory, the deep neural network is highly nonlinear, the TTP of single-track railways are diverse, so we don't recommend using a unified VAF to estimate the state value for various rail lines in different situations, while it is reasonable to use the deep neural network to solve TTP for similar line planning schemes or infrastructures. As shown in Algorithm 6, there is a potential to consider the VAF as an evaluator with offline training experience and to combine it with the online solving made by the planner. For one thing, the fast inference of the learner improves the disadvantage that the planner requires long-time online computing without prior knowledge. For another thing, the planner tends to solve a specific problem instance online, which compensates for the tendency of the learner to apply the commonness of problem instances but no instance-specific features.

## IV. Results

Many studies have been carried out with the assumption that MCTS has an adequate computing budget, such as the application of AlphaGo, but a large amount of computing resources is often hard to obtain for many researchers. Therefore, in order to verify the validity and generality of the proposed methods, we use some random small-scale TTP instances on a single-track rail corridor (for both passenger and freight railways) in a peak period in China. However, when we validate the algorithms with the CSAV strategy, which is designed to deal with large-scale TTP problems that algorithms cannot directly find a feasible solution with high probability under a finite computing budget, we use the medium-scale and real-world instances on the corridor. All experiments are performed on a personal computer equipped with 64G memory and a Nvidia RTX TITAN V GPU.

The selected corridor is a part of Jiaozuo-Liuzhou Railway in Guangxi Province, China (Fig. 13), and the configuration of the corridor is shown in Table 3. The data is supported by the National Railway Train Timetable Research and Training Center of China.

This experiment sets the $\tau_{no}$ and $\tau_d$ of all problem instances to 2 minutes, the $t_{i,j}^{acc}$ and $t_{i,j}^{dec}$ to 1 minute, and the $V_{max}$ to 59. The constant $T'_{travel}$ is set to $V_{max} \times |V|$ for all instances.

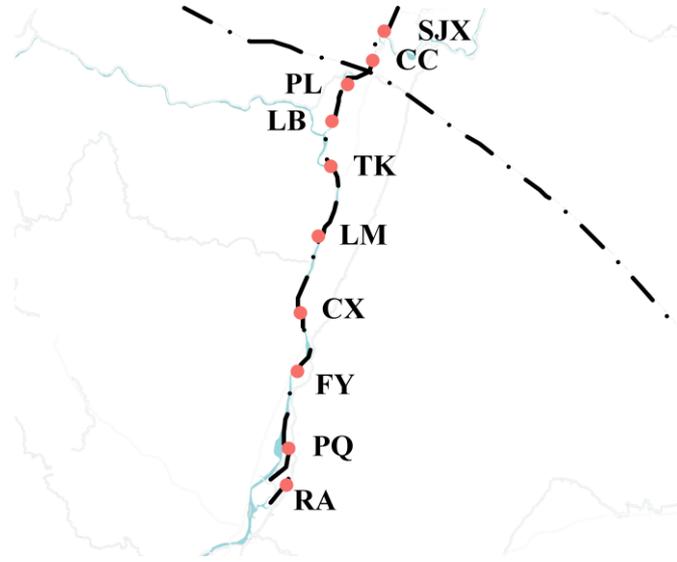

Fig. 13 Diagram of the railway line used in the experiment

Table 3
Line Configuration

| Station name | Number of tracks in station | Section | Section length (meters) |
| --- | --- | --- | --- |
| $(SanJiangXian)\ SJX$ | 3 | $s(SJX, CC)$ | 10300 |
| $(ChengCun)\ CC$ | 3 | $s(CC, PL)$ | 7500 |
| $(PingLiao)\ PL$ | 3 | $s(PL, LB)$ | 6700 |
| $(LaoBao)\ LB$ | 3 | $s(LB, TK)$ | 7300 |
| $(TangKu)\ TK$ | 3 | $s(TK, LM)$ | 10300 |
| $(LiuMeng)\ LM$ | 3 | $s(LM, CX)$ | 11100 |
| $(ChengXiang)\ CX$ | 3 | $s(CX, FY)$ | 9000 |
| $(FuYong)\ FY$ | 3 | $s(FY, PQ)$ | 10200 |
| $(PingQin)\ PQ$ | 3 | $s(PQ, RA)$ | 8100 |
| $(RongAn)\ RA$ | 7 | \ | \ |

## A. Using general UCT algorithms to solve TTP instances

To verify and compare the effectiveness of proposed UCT, UCT_MAX1, and UCT_MAX2 algorithms in a general way, one or more randomly generated instances are used in this part of the experiment. Because these algorithms, theoretically, suppose that the computing budget is adequate, to put it succinctly, the instances with total of 2 to 8 passenger and freight trains are taken to reduce the computing budget requirements. Among the instances, the section running speed of passenger trains is 1250 meters per minute (75km/h), and that of freight trains is 833.3 meters per minute (50km/h). The balance factor $Cp$ is set to 0.7 for all random instances, and the maximum simulation time $T_{max}$ is set to 800.

Comparation of general UCT algorithms in the planner to solve TTP instances

In this paper, 50 TTP instances with different line planning schemes are randomly generated on the railway line for testing the UCT, UCT_MAX1 and UCT_MAX2. We give the same computing budget $CB$ with a value of $10^4$ to the three algorithms, The first departure and destination stations of each train in these instances are randomly generated, so does the type of each train. For different TTP instances, the length of dwell time sequence (corresponding the action sequence) required for decision making is different, so according to the sequence length, 50 instances are sorted for the convenience of comparison. The total train travel time obtained by the three algorithms is indicated by Fig. 14, where the horizontal and vertical axis stand for different instances and the total train travel time.

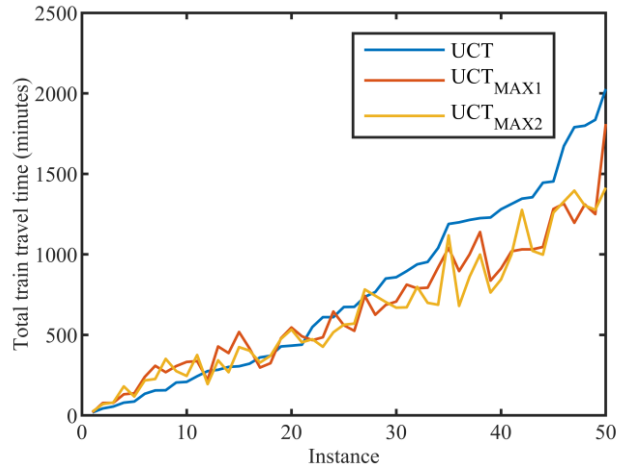

Fig. 14 The total train travel time obtained by the three algorithms on 50 problem instances. The blue, red and yellow polylines represent the UCT, the UCT_MAX1, and the UCT_MAX2 algorithms, in that order. In some instances, the total train travel time may be shorter because the distance between the first departure station of a train and its destination station is close.

From the Fig. 14, UCT_MAX1 and UCT_MAX2 achieve similar total train travel time of feasible solutions with relatively adequate budgets, and as the size of problem instances increases, both algorithms gradually outperform the UCT algorithm. The experimental result first shows that the general UCT algorithms are suitable for solving TTP. Second, with the same computing budget $CB$ and *Simulation* phases of the three algorithms, the result implies that when the instances become more complex, the Max operator enables more exploration of potential nodes, and finally obtains a better feasible solution than UCT. The total train travel time obtained by UCT_MAX1 and UCT_MAX2 in these 50 instances is 13.6% and 17.3% lower than that obtained by UCT, respectively, indicating that the improved algorithms with the Max operator is effective.

Comparation of the general UCT algorithms in the planner with different computing budgets to solve a TTP instance

The performance of three algorithms in different computing budgets is compared to solve a random problem instance, starting with $CB = 10,000$, with an additional $10,000$ simulations each. The total train travel time of three algorithms for solving a feasible solution of the same problem instance is demonstrated in Fig. 15.

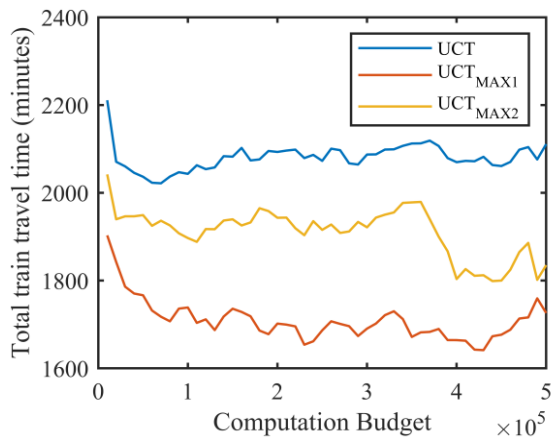

Fig. 15 Comparison of the objective value obtained by the three algorithms with the gradual increase of the computing budget. To show the overall trend, the curves have preprocessed with a moving average function with the window width of 10.

This experiment reveals the difference between UCT_MAX1 and UCT_MAX2. As we can see from Fig. 15, when the computing budget grows, the obtained objective value of UCT_MAX2 is always higher than that of UCT_MAX1. The two algorithms only have a difference in the evaluation functions for nodes, which indicates that using Max operator to guide the computing budget allocation of the algorithm, as shown in Eq. 27, may cause the algorithm to embrace a higher variance. That is to say, when there is a local optimal solution, UCT_MAX2 is more likely to fall into the local optimal solution. For UCT_MAX1, it still uses the average value in the evaluation function, so the interference caused by the occasional local optimization may be weakened by other simulation rewards from the estimated node. Nevertheless, in theory, when an algorithm has enough computing budget, computing resources are inclined to be allocated to the truly optimal nodes, and the interference would be gradually erased.

In addition, it can be concluded from the experimental results that the objective values of the solutions of the three algorithms tend to decline gradually with the increase of the computing budget, but they are not strictly monotonous, which is related to the fluctuations prompted by the randomness of the algorithms. For visualization, the first and 50th timetables obtained with the three algorithms are displayed in Fig. 16.

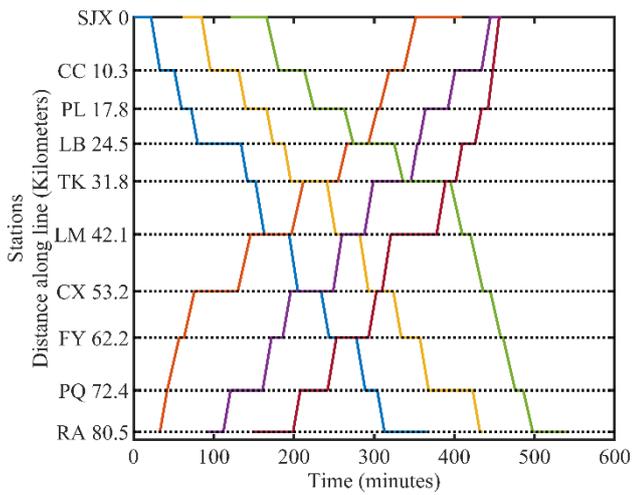

(a)

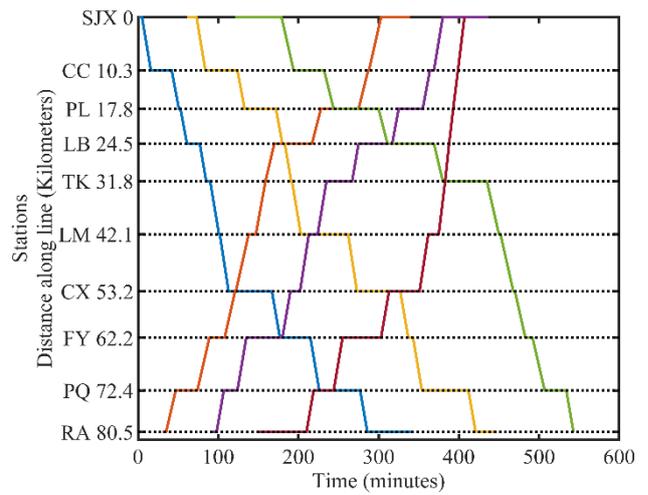

(b)

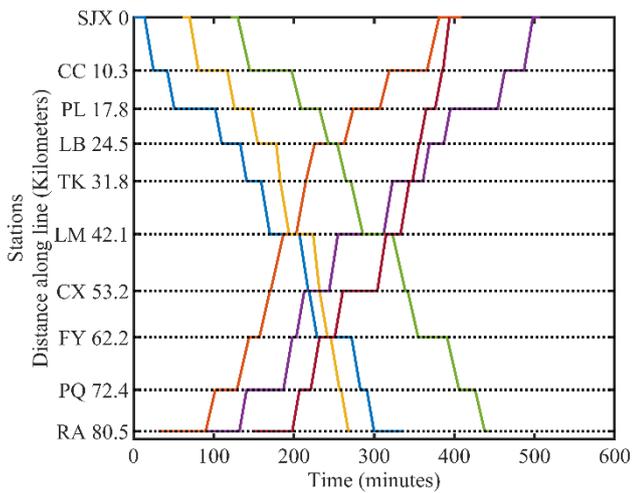

(c)

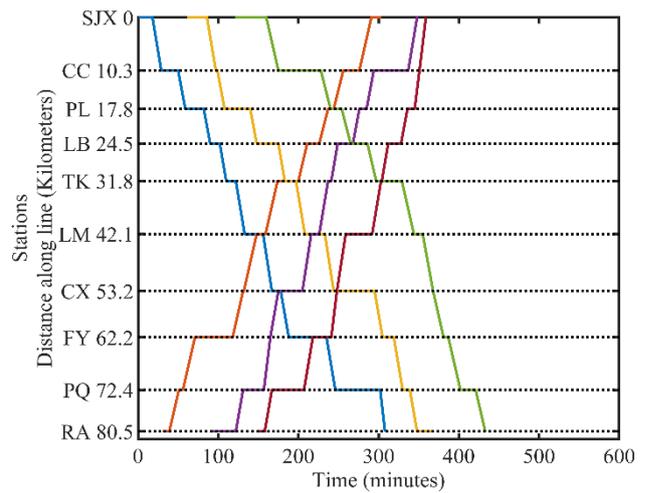

(d)

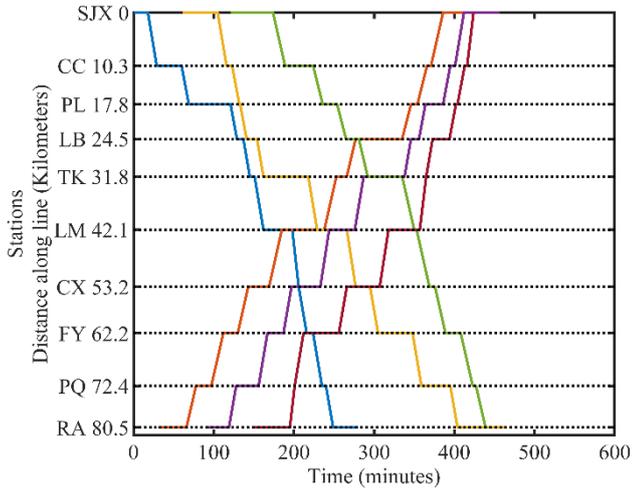
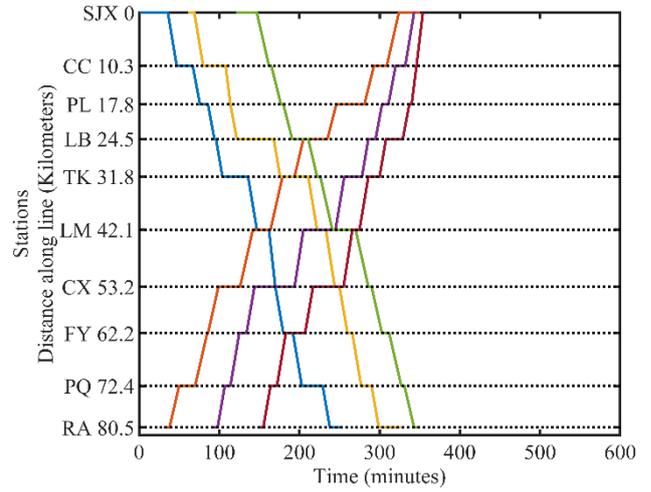

(e)　　　　　　　　　　　　　　　　　　(f)

Fig. 16 The train timetables obtained by the three algorithms when solving a TTP instance for the first time and the 50th time. (a) is the timetable obtained by UCT algorithm in the first computing, with the objective value of 2211, and (b) is the timetable obtained in the 50th computing, with the objective value of 2062; (c) is the timetable obtained by UCT_MAX1 in the first computing, with the objective value of 1903, and (d) is the timetable obtained in the 50th computing, with the objective value of 1666. (e) is the timetable obtained by UCT_MAX2 in the first computing, with the objective value of 2042, and (f) is the timetable obtained for the 50th time, with the objective value of 1520.

In each timetable, the horizontal axis represents the time, and the vertical axis stands for the stations and their mileages from the station $SJX$. Every sideways line indicates that a train is running in a section, where the horizontal part represents the dwell time of the train at the station. The problem instances in this experiment are the same, which are 6 trains running between the $SJX$ station and the $RA$ station. The total train travel time obtained by UCT, UCT_MAX1 and UCT_MAX2 decreases by about 5.5%, 9.3% and 10.1%, respectively, compared with the first time, with the increase of the computing budget. The result strengthens the asymptotic optimality of UCT algorithm in theory.

## B. Using the general UCT algorithms with CSAV to solve an instance in the case of insufficient computing budget

Theoretically speaking, with sufficient computing budget, UCT, UCT_MAX1 and UCT_MAX2 can obtain optimal solutions of problem instances, but as a matter of fact, when the problem instance is larger, many algorithms fail to obtain an optimal solution in an acceptable time. In order to test UCT_CSAV, the general UCT algorithms combining with CSAV, we set 120 random instances, and each instance contains 30 passenger and freight trains in total with different first departure time. The computing budget is set to 500,000 for all algorithm configurations. Since CSAV could combine with different algorithms within the general UCT algorithms, we test all general UCT algorithms integrating with the CSAV. In these experiments, the weighting coefficient $\alpha$ and the $\beta$ described in Eq. 30 are set to $10^{-3}$ and $2.5 \times 10^{-4}$, and the weighting parameter $Cp$ in UCB (Eq. 23) is set to 0.1. The maximum simulation time $T_{max}$ is specified as 1800. When the computing budget $CB$ is $5 \times 10^5$, the number of feasible solutions that can be obtained by each algorithm configuration and the average total train travel time of feasible solutions are indicated by Table 4.

Table 4 The number of feasible solutions that can be obtained by different algorithm configurations and the average total train travel time.

| Algorithm Configuration | Average total train travel time | The number of feasible train timetables that each algorithm can obtain in 120 instances within the computing budget |
|---|---|---|
| General UCT algorithms | | 0 |
| UCT_CSAV (UCT with CSAV) | 13165 | 20 |
| UCT_CSAV (UCT_MAX1 with CSAV) | 12549 | 96 |
| UCT_CSAV (UCT_MAX2 with CSAV) | 11823 | 51 |

In these TTP instances, the length of the action sequence is 300, that is, the number of actions in the action space is about $10^{533}$, implying that the computing budget is relatively insufficient. As shown in Table 4, the general UCT algorithms cannot obtain a feasible solution, but algorithms combining CSAV can. Besides, the UCT_MAX1 integrating with CSAV is able to get the most feasible solution, but its quality is lower than that of UCT_MAX2 combining with CSAV. To further explain how the CSAV works, Fig. 17 presents a diagram of the trains (the black line) being decided by the algorithm and the subsequent trains (green lines) that the CSAV can evaluate when the UCT_MAX2 integrating with CSAV solves an instance. Although the UCT_MAX2 could not obtain a complete and feasible solution in the early planning stage, heuristics still provide a reasonable estimate of the action value of the UCT_CSAV based on the potential timetables of some subsequent trains.

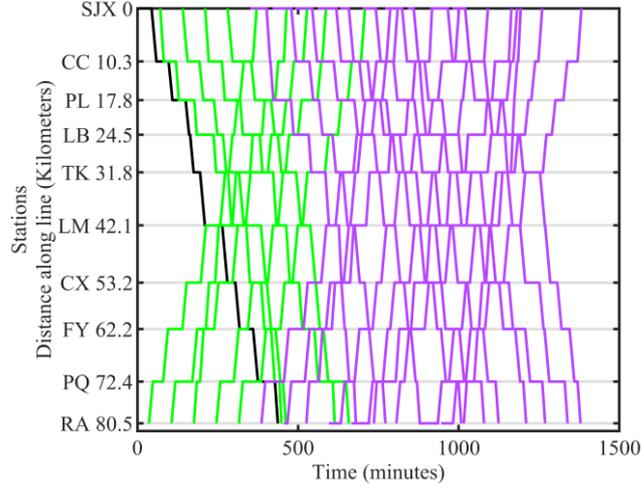

Fig. 17 A train timetable with a total train travel time of 11548. The timetable is obtained from a random instance by the UCT_MAX2 combining with CSAV. By the CSAV, the train timetable represented by the green lines provides a value estimation for the decision of the train represented by the black line.

When computing resources are sufficient, in theory, the purple lines would be replaced by green lines. Therefore, the performance of general UCT algorithms with and without the CSAV is the same. In other words, the asymptotic optimality of the general UCT algorithms is unaffected under the circumstance of sufficient computing budgets.

**C. The learner learns the knowledge obtained by the planner**

First, we use the UCT algorithm with a computing budget of 10000 to solve 320 random instances, and collect about 530000 state-value pair training data to generate the training dataset $DS$. In data preprocessing, the width and height of the Executing window is set to 121 and 43 (the scale can be adjusted based on computing abilities). The size of each focused local matrix is $3 \times 43 \times 121$, and the input size of the entire matrix is the size of the simulation matrix, $3 \times 43 \times 800$. The number 3 in the size indicates that the input state is stacked with three preorder node states, and the input states are similar to the composition of RGB images, so that the input can be visualized in the form of images.

Then, we instantiate a VAF, and its deep neural network structure is presented in Table 5. By comparison, it can be found that using Dropout technology can improve the robustness and generalization ability of the network, so the Dropout operation is added to the VAF before the output layer. Also, a state-value pair is set as $st, Q(st)$. According to the supervised learning method, the target of network parameter update is to minimize the difference $\delta$ between $f_{vaf}(st)$ and $Q(st)$. In this paper, the smooth L1 function [58] is utilized to estimate the difference between the two (Eq. 31), and Adam [59] with the learning rate of 0.001 is used to optimize the network parameters. More than that, we use the L1 parameter regularization technique, which is designed to reduce the interference of irrelevant neurons on the results, to further strengthen the generalization of VAF. The error $\bar{\delta}$ in the actual training is shown in Eq. 32, but the error in the inference remains $\delta$.

Table 5 Parameters of VAF

| Item | | Value |
|---|---|---|
| Convolutional layer 1 | Number of kernels | 64 |
| | Size of the kernel | $5 \times 5$ |
| | Stride step | 3 |
| Convolutional layer 2 | Number of kernels | 128 |
| | Size of the kernel | $3 \times 3$ |
| | Stride step | 2 |
| Convolutional layer 3 | Number of kernels | 64 |
| | Size of the kernel | $3 \times 3$ |
| | Stride step | 1 |
| Convolutional layer 4 | Number of kernels | 128 |
| | Size of the kernel | $3 \times 3$ |
| | Stride step | 2 |
| Full connection layer 1 | Number of neurons | 101376 |
| Full connection layer 2 | Number of neurons | 512 |
| Full connection layer 3 | Number of neurons | 151040 |
| Full connection layer 4 | Number of neurons | 256 (*For connections with Full connection layer* 2) + 128 (*For connections with Full connection layer* 3) |
| Dropout layer | Proportion of neurons ignored | 0.5 |
| | Number of neurons | 128 |
| Full connection layer 5 | Number of neurons | 1 |

$$\delta = \begin{cases} 0.5(f_{vaf}(st) - Q(st))^2, & if\ |f_{vaf}(st) - Q(st)| < 1 \\ |f_{vaf}(st) - Q(st)| - 0.5, & otherwise \end{cases} \quad (31)$$

$$\overline{\delta} = \delta + \theta \sum_{w \in vaf} |w| \quad (32)$$

$w$ is the neural weight parameter in the deep neural network VAF, and $\theta$ is the weighting factor, $\theta = 10^{-5}$.

Finally, the performance is tested on the testing dataset by various VAF configurations (as shown in Fig. 18), including the VAF
- using the MSE function $(f_{vaf}(st) - Q(st))^2$ or the smooth L1 function as the error function;
- with/without the Dropout technology;
- with/without the L1 parameter regularization technology;
- with/without doubling the original dataset by adding new similar training data, respectively.

Each configured VAF reads 256 training data each Epoch for batch training. The VAF marked with MSE in Fig. 18 indicates that its error function uses $(f_{vaf}(st) - Q(st))^2$. When the MSE is used as the error function, the algorithm tends to diverge on the testing dataset. Through checking its network, it is found that the gradient is too large, but this phenomenon is improved after using smooth L1 error function and Dropout technology. However, as the training progresses, VAF without L1 regularization technology has different degrees of over fitting, resulting in poor generalization ability of the algorithm on the testing dataset. After adding L1 regularization, the over fitting is suppressed. The reason for this situation is that many irrelevant neurons interfere with the generalization of the algorithm owing to the large network scale of VAF, and after regularization, the weight $w \in vaf$ of irrelevant neurons gradually decreases, which improves the over fitting. When VAF gets twice as much the data for training, its generalization ability is significantly enhanced. Fortunately, as mentioned above, the MCTS framework proposed in this paper can generate a large number of state-value pairs by itself, and the data quality improves with the increase of computing budget $CB$, signifying that the VAF is capable of getting a lot of training offline and making full use of existing computing resources.

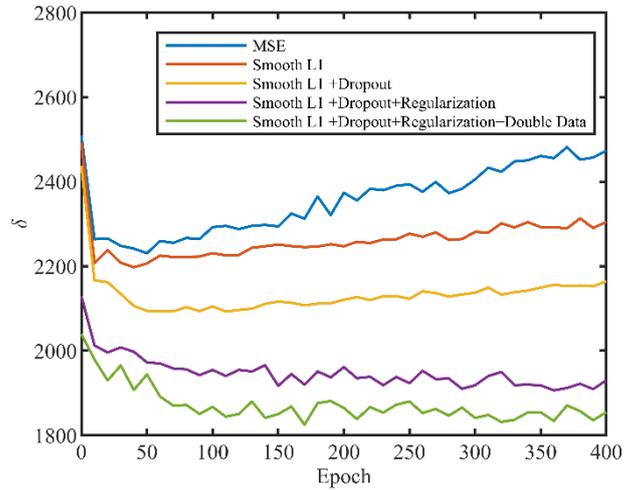

Fig. 18 Performance comparison of various configured algorithms on the testing dataset. The testing dataset is independent of the training dataset. The instance number of testing dataset is one-fourth of the training data, and the testing dataset has not been trained in VAF.

The performance of VAF on the testing dataset shows that it has the ability to learn the input multi-layer state matrix data, and has a certain generalization ability on the unlearned data. As mentioned earlier, the state $st$ is a 3-layer and 2-dimensional matrix, which is convenient for visualization and VAF using CNN. This also signals that there is an opportunity for more new advances in computer vision to be applied to solve TTP.

Deep convolutional networks, in essence, process the state $st$ by extracting the local and regional features and identifying them to estimate the value of the state $Q(st)$. The features of the state extracted from the convolutional layers in the VAF are shown in Fig. 19. It can be seen from Fig. 19 that the feature maps of the focused local input matrix have more details, which is helpful for the VAF to identify the information most relevant to the train being decided. After multiple convolutional layers, the number of pixels in the image is less, which is conducive to reducing the required neuron counts to process the entire image input. In addition to the strong ability to extract regional features, reducing the computing cost is also the reason why we choose the CNN.

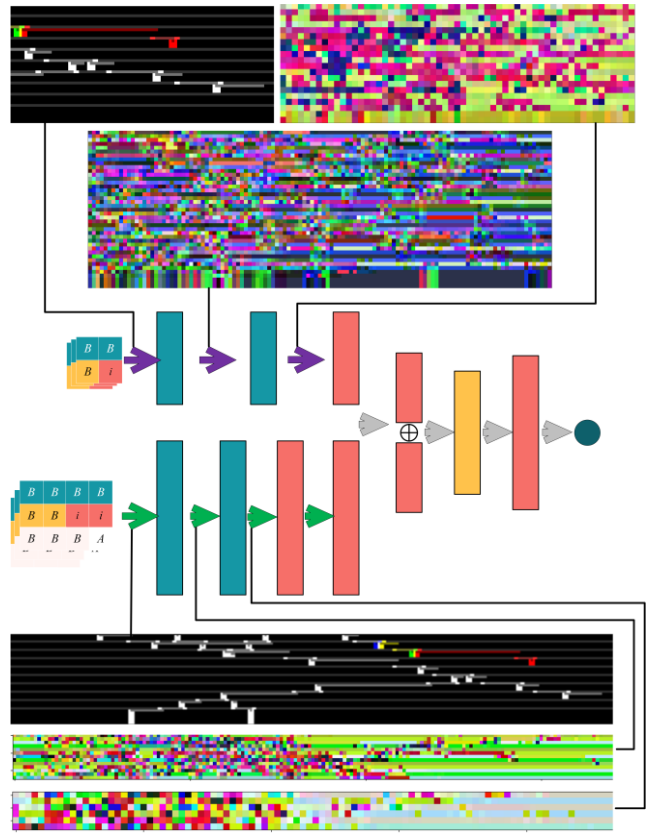

Fig. 19 Feature maps of a state data extracted from the VAF. The images in the upper part are the focused local input matrix image and its generated feature maps by the first and second convolutional layer respectively. The lower part is the image of the entire input matrix and its generated feature maps by the first and second convolutional layer respectively. The corresponding positions of each image in the network is connected.

### D. Using the integrated framework with the planner and learner to solve random and real-world instances

The planner has various algorithm configurations, and we use the simplest one, UCT, in the planner to combine with the VAF in the learner, which is recorded as UCT_VAF. Its effectiveness is verified on 50 random TTP instances. When dealing with a large-scale and real-world TTP instance, the UCT_MAX2 algorithm with the CSAV in the planner is adopted to combine with the VAF in the learner, which is expressed as UCT_CSAV_VAF. We test it in a peak period on a railway corridor, and compare it with a RL algorithm and a commercial solver.

Solving the random instances

Considering a simple algorithm configuration in planner and learner, we embed the trained VAF described in Sec. IV-C into the planer with the UCT to instantiate an integrated algorithm UCT_VAF. In order to test this algorithm, another 50 random TTP instances are generated and solved by UCT and UCT_VAF respectively. The maximum simulation time $T_{max}$ is set to 800 in this experiment. The comparison results are shown in Fig. 20.

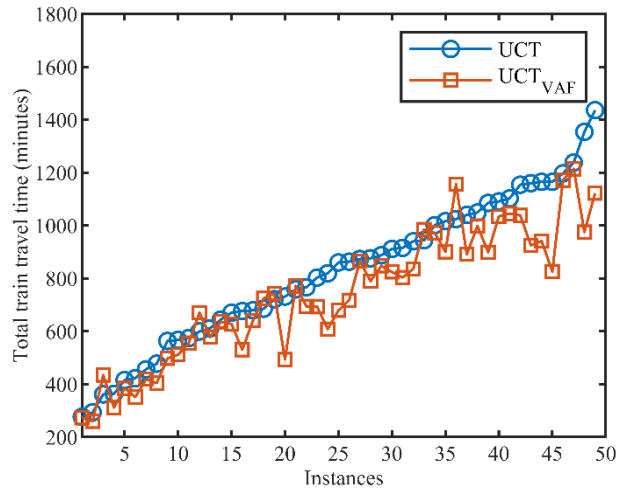

Fig. 20 Performance of UCT and UCT_VAF in 50 random instances.

As shown in Fig. 20, the UCT_VAF outperforms than the UCT in most instances. Since the configuration and computing budget $CB$ of the two algorithms are both 10000, we believe that the node Warm up method shown in Algorithm 6 makes potential child nodes get more explorations. Compared with the computing time of the UCT itself, the time that the trained VAF only infers the value of the child nodes of the selected node in the *Expansion* phase is negligible. However, the pre-training process of VAF requires a lot of computations, this offline training and online planning method proposed in this paper can make full use of the existing computing resources, especially for railway enterprises. This advantage is not available in many current algorithms, including the RL algorithm. Moreover, according to the experimental results of Sec. IV-A and Sec. IV-C, when both the computing budget of the algorithm in the planner and the number of training samples for VAF increases, the performance of the integrated framework will continue to improve.

It is not cost-effective to use one unified VAF to solve the TTP under various railway line environments or line planning schemes. Therefore, for the tested UCT_VAF method, the problem instances are limited to a railway corridor, so as to reduce the diversity of problems.

Solving the real-world instances

In order to simulate the corridor in a peak period, we assume 52 passenger and freight trains in total to test the integrated framework. This experiment instantiates a VAF with the same structure as described in Sec. IV-C, but the training data is generated by UCT_CSAV, as shown in Sec. IV-B. In offline-training and online-solving stages, since the input data size is larger than the size required by the instantiated VAF, it is cut in the data pre-processing according to the location of the current decision-making train in the state to make the input size meet the requirements of VAF. That is, the size of each focused local matrix is $3 \times 43 \times 121$, and the size of the entire matrix is $3 \times 43 \times 800$. After the training is completed, the UCT_CSAV_VAF is used to solve the real-world instances, and the train timetable is obtained in Fig. 21.

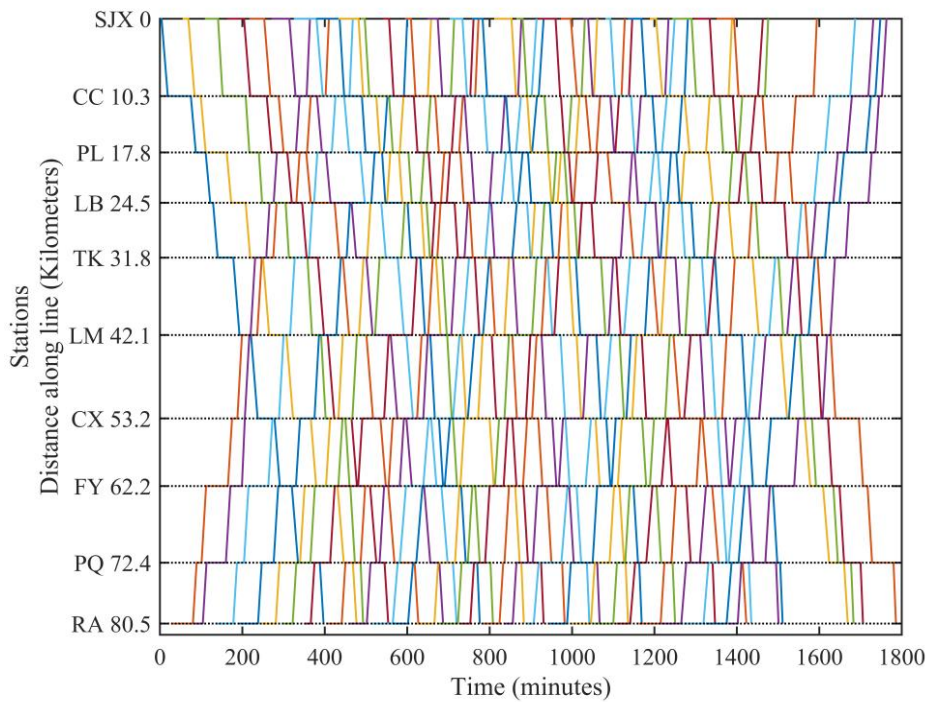

Fig. 21 The train timetable obtained by UCT_CSAV_VAF for the instances. A colored line stands for a train.

Different from the train timetable obtained in the previous experiments, this train timetable is greatly affected by the station capacity constraints. For many stations, keeping 2 or 3 trains dwelling at the same time for a long period reaches their capacity limit. This not only shows that the station capacity constraint should not be ignored, but also reflects that the integrated method can solve the feasible train timetable under the condition of the fully limited station capacity. It can be seen that, although the scale of instances used for training is smaller, the VAF still plays a role in estimating the value of the state matrix in the larger instance. This may attribute to the generalization brought by the setting of focused local matrix input of VAF, because the features of the input focused local matrix in this instance and in the instances for training are consistent.

In order to compare the effectiveness of the integrated framework, we test the three algorithms, UCT, UCT_MAX1 and UCT_MAX2. With the same computing budget, UCT_CSAV_VAF always has the capability to find the feasible solution first. Also, the performance of the ASRL algorithm[24] and IBM CPLEX Optimizer in this instance is tested. The ASRL is a scalable RL algorithm for TTP based on the capacity information of the line around the train being decided. However, due to the limited ability of perceiving the time constraint information of the ASRL, this algorithm ends in failure after solving the conflict-free timetable of about 10 trains in this instance. For CPLEX Optimizer, we input the instance in the form of MIP model as described in the Sec-II, and CPLEX cannot find a feasible solution within 48 hours. By comparison, the effectiveness of our proposed method is further highlighted. Another advantage is that the quality of the solution is clearly related to the computing budget, which helps railway enterprises to evaluate the computing resources to be allocated according to the instances required to be solved.

# V. Conclusion

Firstly, we build an MIP model for the TTP of single-track railways, and design a state matrix to express the train trajectory and constraint distribution during train operations. Based on the representation method of this state matrix, a train operations simulator is designed, and the dynamics model of the simulator is given.

According to the designed simulation method, an algorithmic framework, consisting of a planner and a learner, is proposed for solving the TTP. Multiple MCTS algorithms in the planner can solve TTP instances and provide a large amount of training data for the learner; and the knowledge learned by the learner can be used to provide a priori approximate value of each state for the algorithms in the planner in order to improve their searching efficiency. The UCT algorithm is regarded as the basic algorithm of the planner, and it can gradually converge to the optimal solution of the TTP instance with the sufficient computing budget. Considering that optimizing train timetables tends to minimize the total travel time of all trains, we propose two improved UCT algorithms, UCT_MAX1 and UCT_MAX2, for solving TTP. Although the solution quality of the general UCT algorithms increases with their computing budgets, the computing budgets required by general UCT algorithms cannot be guaranteed when considering large-scale TTP instances. As a matter of fact, the computing budgets they have are far from enough to find a feasible solution. To address this issue, a heuristic strategy CSAV in the planner is put forward to improve the searching efficiency of the general UCT algorithms without affecting their theoretical asymptotic optimality, so that users can easily configure the desired combination of algorithms in the planner according to the scale of the TTP instances and the computing resources they have.

Since the MCTS algorithms in the planner are based on the designed simulator, many state matrixes and its estimated values are generated in the simulation process. To learn the relationship between the two in these data, the learner uses a deep CNN to learn these data, and the network is treated as a VAF to infer the value of the input state matrix. The experiment shows that the deep convolutional network of the learner has the ability to learn and generalize the state matrix data. Therefore, we propose a solving framework for TTP that integrates a planner and a learner. That is to say, the trained VAF provides a priori value estimation for some state matrixes reached by the algorithms in the planner, and guide the algorithms to search more potential nodes. The experiment indicates that this integrated framework is more effective in solving TTP instances than the planner alone.

In summary, the planner and the learner in the proposed method are mutually beneficial. On one hand, when the planner gets more computing budgets, the training data will be more accurate, which is good for the learner to estimate the value of the state matrix more accurately. Once the accuracy of the learner improved, the planner would obtain a better searching efficiency, and the quality of the solution is enhanced. What worth mentioning is that this integrated framework can generate a large number of labeled training data by itself, saving the data annotation cost compared with it in the common supervised learning. More than that, the offline training approach can make full use of the existing computing resources from railway enterprises and continuously improve the effectiveness of the integrated framework. Thus, this integrated framework is promising in providing a new application paradigm for offline-training and online-solving schemes for TTP research using artificial intelligence methods. On the other hand, in general, the knowledge of state value estimation obtained by conventional MCTS algorithms in the simulation process cannot be reused, which reduces the computing efficiency of the algorithms. However, the learner can fully utilize the knowledge and feed it back to the MCTS algorithms in the planner, achieving the effect of "one plus one is greater than two".

Furthermore, when computing resources are insufficient and feasible solutions cannot be obtained, it is important for a planner to estimate the value of the reached state matrixes. Although we propose a heuristic method to strengthen the searching efficiency of planners, a better estimation method deserves further research. Inspired by the relevant research of AlphaGo and its lead author, David Silver, the random simulation action selection strategy of the MCTS algorithms in the planner may be improved, such as using deep RL algorithms (like DQN[60]) for simulation action selection, which is also one of the research interests that our team is focusing on. Moreover, from the perspective of application, it is also an interesting research direction to apply the proposed integrated framework to the related fields such as energy-efficient and robust train timetable optimization.